\documentclass[sigconf,authorversion,nonacm]{aamas} 

\usepackage{balance} 

\DeclareMathOperator{\E}{\mathbb{E}}

\def\Var{{\textrm{Var}}\,}
\def\max{{\textrm{max}}\,}

\usepackage{subcaption}

\usepackage{algorithm} 
\usepackage{algorithmic}

\title[AAMAS-2024 Formatting Instructions]{Guided Exploration in Reinforcement Learning \\ via Monte Carlo Critic Optimization}

\author{Igor Kuznetsov}
\affiliation{
  \institution{Independent Researcher}
  \city{Oxford, UK}
  \country{United Kingdom}}
\email{igorkuznetsov14@gmail.com}

\begin{abstract}
The class of deep deterministic off-policy algorithms is effectively applied to solve challenging continuous control problems. Current approaches commonly utilize random noise as an exploration method, which has several drawbacks, including the need for manual adjustment for a given task and the absence of exploratory calibration during the training process. We address these challenges by proposing a novel guided exploration method that uses an ensemble of Monte Carlo Critics for calculating exploratory action correction. The proposed method enhances the traditional exploration scheme by dynamically adjusting exploration. Subsequently, we present a novel algorithm that leverages the proposed exploratory module for both policy and critic modification. The presented algorithm demonstrates superior performance compared to modern reinforcement learning algorithms across a variety of problems in the DMControl suite.
\end{abstract}


\newcommand{\BibTeX}{\rm B\kern-.05em{\sc i\kern-.025em b}\kern-.08em\TeX}

\begin{document}


\pagestyle{fancy}
\fancyhead{}


\maketitle 


\section{Introduction}

In reinforcement learning, exploration plays a crucial role in policy optimization. The chosen exploration strategy defines the interaction with the world, thereby influencing the ultimate success of the agent and proving pivotal in solving continuous problems \citep{pmlr-v139-zhang21t}, navigation challenges \citep{mazoure2020leveraging}, and hierarchical issues \citep{vigorito2016intrinsically}. At a higher level, exploration methods can be broadly categorized as undirected or directed \citep{thrun1992efficient}. Undirected methods involve generating random exploratory actions based on a desired exploration-exploitation trade-off, while directed algorithms rely on information provided by a policy or a learned world model.

\begin{figure}[h]
  \centering
  \includegraphics[width=1.0\linewidth]{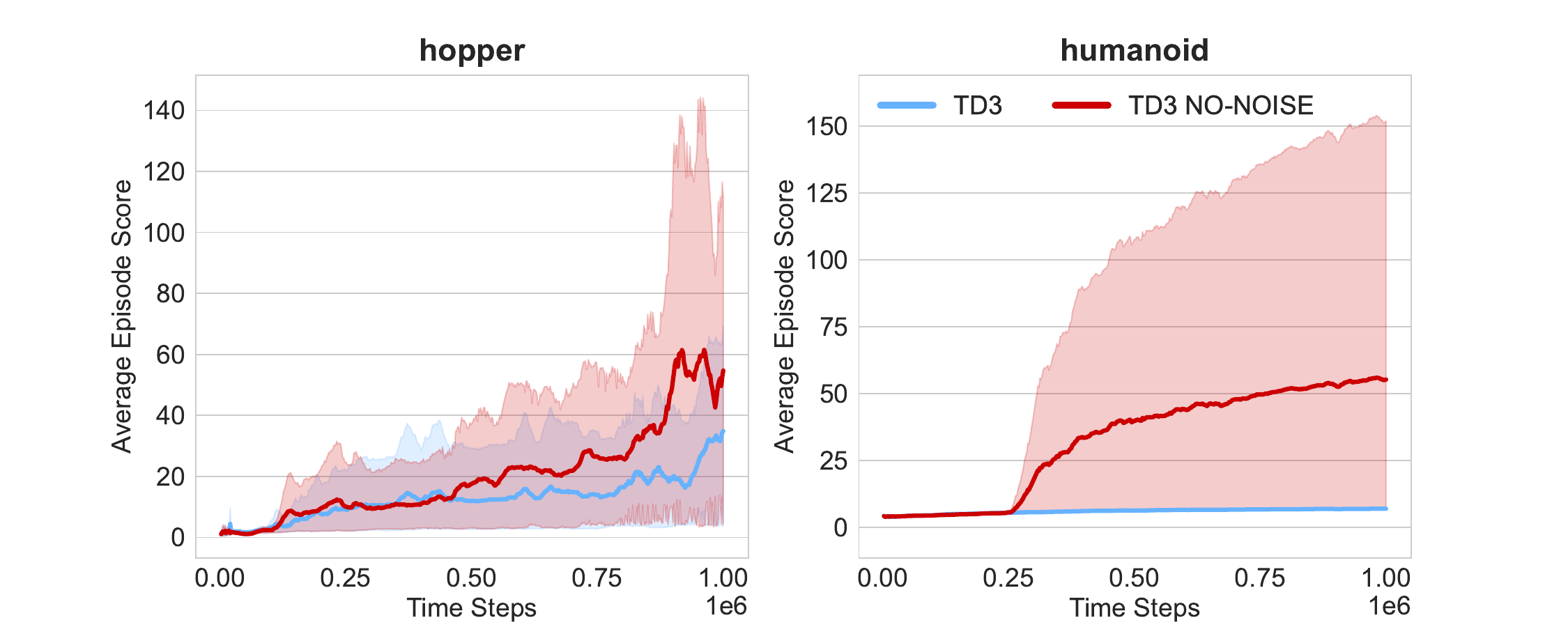}
  \caption{Preliminary motivation experiment. The performance of the original version of TD3 algorithm compared with a variant without exploration noise.}
  \label{fig:ge_scheme}
  \Description{Preliminary motivation experiment.}
\end{figure}

In the real world, humans and animals seldom rely on random exploration of the environment, consistently pursuing intrinsic motivations such as safety \citep{tully2017exploration}, curiosity \citep{berlyne1966curiosity, kidd2015psychology}, or uncertainty minimization \citep{gershman2019uncertainty}. Equivalents of these psychological phenomena have been successfully applied within the corresponding research directions of reinforcement learning, forming the class of intrinsically motivated algorithms \citep{garcia2015comprehensive, pathak2017curiosity}. A schematic representation of such intrinsically motivating guided exploration is depicted in Figure~\ref{fig:ge_scheme}, illustrating different directions of policy parameter changes for external- and intrinsic-reward objectives.

\begin{figure}[t]
  \centering
  \includegraphics[width=0.8\linewidth]{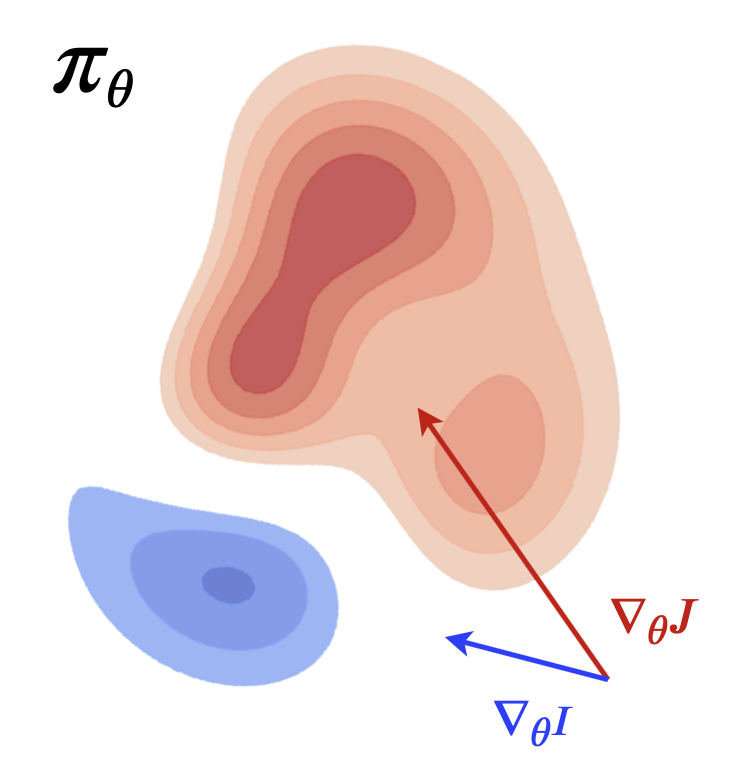}
  \caption{A scheme of a guided exploration model. Policy gradients $\nabla_\theta$  are directed by the final reward objective $J$ and corrected by an intrinsic objective $I$ facilitating directed exploration.}
  \label{fig:no_noise}
  \Description{A scheme of a guided exploration model.}
\end{figure}

In the context of continuous control problems, the class of deep deterministic off-policy methods has gained high popularity in the reinforcement learning community due to its implementation simplicity and state-of-the-art results. However, from an exploration perspective, algorithms such as DDPG \citep{lillicrap2015continuous} and TD3 \citep{fujimoto2018addressing} utilize Gaussian or time-dependent Ornstein-Uhlenbeck noise applied to deterministic actions. While serving as a simple and effective exploration tool, methods based on random noise face three limitations. Firstly, in high-dimensional environments, random exploration proves to be an inefficient strategy for reaching specific regions of interest in the state-space that are critical for success \citep{burda2018exploration}. Secondly, the discussed algorithms apply noise with the same mean and deviation throughout the entire period of policy optimization. While crucial in the initial stages of parameter search, excessive exploration may hinder the policy's performance in reaching its optimal condition. Finally, the previously mentioned methods apply noise of uniform magnitude to all actions, disregarding the fact that different components of the action vector in continuous control frequently yield varying magnitudes.

To illustrate the sub-optimality of conventional random exploration and justify our motivation, we evaluate the performance of the TD3 algorithm on two continuous control domains: \textit{hopper-stand} and \textit{humanoid-stand}. We compare the original version of TD3 with a variant that excludes any noise during action selection. The original TD3 exhibits poor performance when contrasted with the noise-free version (Figure~\ref{fig:no_noise}). This straightforward experiment highlights the detrimental impact of current Gaussian-based exploration on the agent's performance. However, it is important to note that the complete absence of exploration may also result in lower episodic rewards, as demonstrated in our subsequent experiments. Our goal is to develop an effective method of directed exploration that enhances the outcomes of deterministic off-policy algorithms.

We address the mentioned weaknesses of random exploration by by introducing a differentiable module capable of guiding a policy in a purposeful exploratory direction. Specifically, we advocate using an ensemble of Q-function approximations trained to predict Monte Carlo Q-values as this controller. Through optimizing multiple independent predictions and utilizing the action gradient derived from calculated variance, we can obtain an uncertainty estimate for a given state. Using this gradient, we introduce an exploratory action correction aimed at exploring the least traversed regions of the environment.

This paper makes a twofold contribution. Firstly, we introduce a guided exploration method that can replace Gaussian noise and seamlessly integrate into any off-policy deterministic algorithm. We demonstrate that our proposed exploration outperforms other exploration methods across a range of continuous control tasks. Secondly, we introduce a novel algorithm that leverages the proposed uncertainty-based module for both the actor and critic components of the model architecture. Our proposed method demonstrates superior results when compared to contemporary reinforcement learning algorithms on a set of tasks from the DMControl suite \citep{tassa2018deepmind}.

Following the guidelines of reproducibility of reinforcement learning algorithms \citep{henderson2018deep} we report the results on a large number of seeds and release the source code alongside raw data of learning curves \footnote{%
  Code is available at \url{https://github.com/schatty/MOCCO}
}.


\section{Preliminaries}

We consider a standard reinforcement learning (RL) setup, in which an agent interacts with an environment $\mathcal{E}$ at discrete time steps aiming to maximize the reward signal. The environment is a Markov Decision Process (MDP) that can be defined as $\langle \mathcal{S}, \mathcal{A}, \mathcal{R}, \rho, \mathcal{\gamma} \rangle$, where $\mathcal{S}$ is a state space, $\mathcal{A}$ is an action space, $\mathcal{R}$ is a reward function, $\rho$ is a transition dynamics and $\gamma \in [0, 1]$ is a discount factor. At time step $t$ the agent receives state $s_t \in \mathcal{S}$ and performs action $a_t \in \mathcal{A}$ according to policy $ \pi$, a distribution of $a$ given $s$ that leads the agent to the next state $s_{t+1}$ according to the transition probability $\rho(s_{t+1}|s_t, a_t)$. After providing the action to $\mathcal{E}$, the agent receives a reward $r_t \sim \mathcal{R}(s_t, a_t)$. The discounted sum of rewards during the episode is defined as a \textit{return} $R_t = \sum_{i=t}^T \gamma^{i-t} r(s_i, a_i)$.

The RL agent aims to find the optimal policy $\pi_\theta$, with parameters $\theta$, which maximizes the expected return from the initial distribution $J(\theta)=\E_{s_i \sim \rho_\pi, a_i \sim \pi_\theta } [R_0]$. The action-value function $Q$ is at core of many RL algorithms and denotes the expected return when performing action $a$ from the state $s$ following the current policy $\pi$:
\begin{eqnarray}
Q^\pi(s, a) = \E_{s_i \sim \rho_\pi , a_i \sim \pi} \left [ R_t | s, a \right].
\end{eqnarray}
In continuous control problems the actions are real-valued and the policy \(\pi_\theta\) can be updated taking the gradient of the expected return \( \nabla_\theta J(\theta) \) with deterministic policy gradient algorithm ~\citep{silver2014deterministic}:
\begin{equation}
    \nabla_\theta J(\theta) = \E_{s \sim \rho_\pi } \left[ \nabla_{a} Q^\pi (s, a) |_{a=\pi(s)} \nabla_{\theta} \pi_\theta (s) \right].
\end{equation}
The class of actor-critic methods operates with two parameterized functions. An actor represents policy \(\pi\) and the critic is the Q-function. The critic is updated with temporal difference learning by iteratively minimizing the Bellman error ~\citep{watkins1992q}:
\begin{equation}
    J_Q = \E \left[ (Q(s_t, a_t) - (r + \gamma Q(s_{t+1}, a_{t+1})))^2 \right].
\end{equation}
In deep reinforcement learning, the parameters of Q-function are modified with additional frozen target network \(Q_{\theta'}\) which is updated by \(\tau\) proportion to match the current Q-function \( \theta^{'} \leftarrow \tau \theta + (1-\tau) \theta^{'} \)
\begin{equation}
    J_Q = \E \left[ (Q(s_t, a_t) - Q')^2 \right],
\end{equation}
where
\begin{equation}
Q'=r(s_t, a_t) + \gamma Q_{\theta'}(s_{t+1}, a'), a' \sim \pi_{\theta^{'}}(s_{t+1}).
\end{equation}
The actor is learned to maximize the current \(Q\) function: 
\begin{equation}
    J_\pi = \E \left[ Q(s, \pi(s)) \right].
\end{equation}
In this work, we focus on an off-policy version of actor-critic algorithms, proven to have better sample complexity \citep{lillicrap2015continuous}. Within this approach, the actor and the critic are updated with samples from a different policy, allowing to re-use the samples collected from the environment. The mini-batches are sampled from the experience replay buffer \citep{lin1992self}. 

In off-policy Q-learning, actions are selected greedily w.r.t. maximum Q-value thus producing overestimated predictions. An effective way to alleviate this issue is keeping two separate Q-function approximators and taking a minimal one during the optimization \citep{van2016deep}. Modern actor-critic methods \citep{fujimoto2018addressing,haarnoja2018soft} independently optimize two critics $Q^1, Q^2$ with identical structure and use the lower estimate during the calculation of the target $Q'$:
\begin{equation}
    Q'=r(s_t, a_t) + \gamma \min{[Q^1_{\theta'}(s_{t+1}, a'), Q^2_{\theta'}(s_{t+1}, a')]}.
\end{equation}


\section{Method of Guided Exploration}

On a high level, our agent consists of two components: an actor-critic part to provide a deterministic policy and an exploratory module that adjusts the policy output action to facilitate directed exploration. 
The deterministic policy $\pi_\theta(s)$ is parameterized by $\theta$ and optimizes reward maximization to produce a \textit{base} action $a^\mathrm{b}$. The exploratory module $\mathrm{EM}$ optimizes auxiliary intrinsic objective to produce an \textit{exploratory} action correction $a^\mathrm{e}$. The policy and the exploratory module are jointly optimized to produce an additive action for collecting transitions for off-policy updates:
\begin{equation}
a \coloneqq a^\mathrm{b} + a^\mathrm{e}, a^\mathrm{b} \sim \pi_\theta(s)
\end{equation}
For the actor-critic part, we use the TD3 algorithm \citep{fujimoto2018addressing} as a backbone.

\begin{figure}[h]
  \centering
  \includegraphics[width=0.75\linewidth]{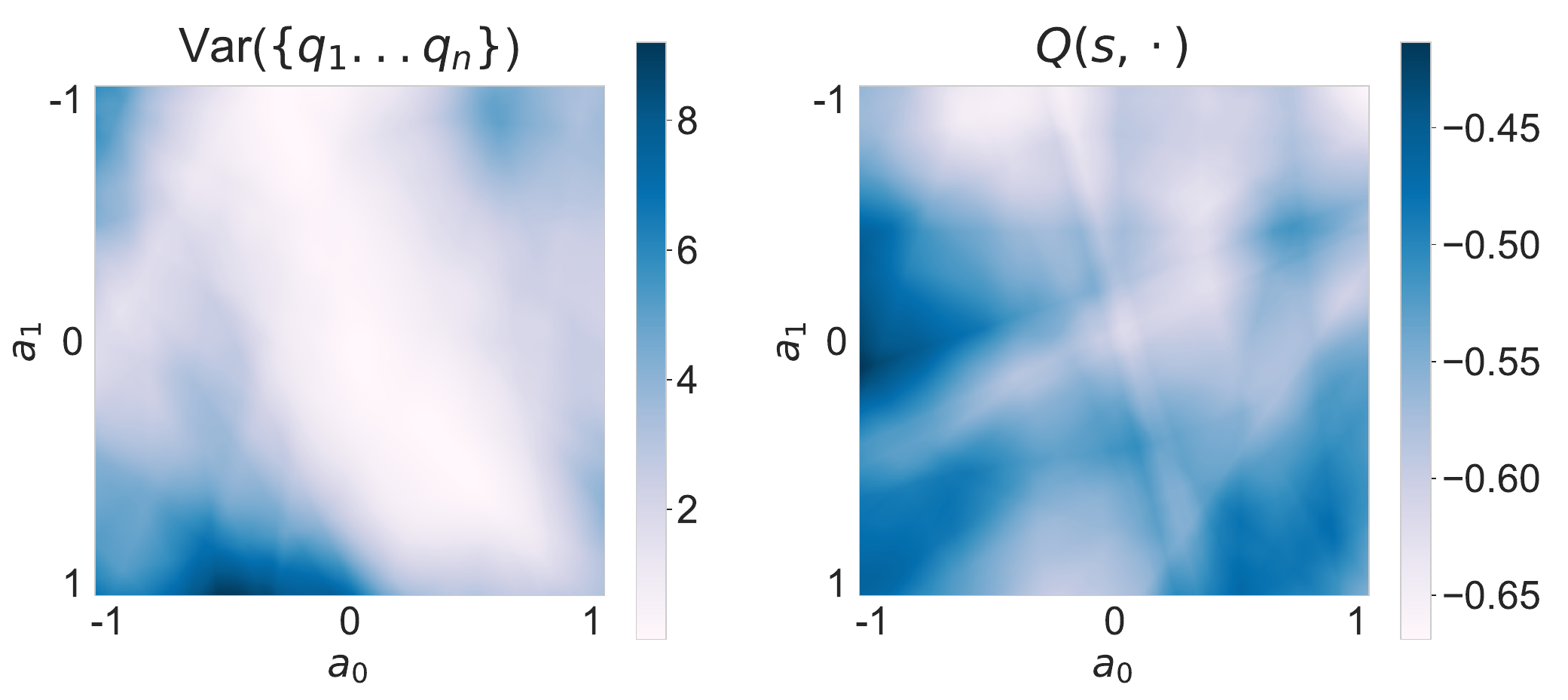}
  \caption{Visualization of uncertainty estimation (left) and critic prediction (right) at the same state on 2D action plane. Environment: point\_mass-easy.}
  \label{fig:viz_uncertainty}
  \Description{Visualization of uncertainty estimation.}
\end{figure}

\subsection{Exploratory Module}
The exploratory module $\mathrm{EM}_\omega(s|\theta)$ is parameterized by $\omega$ and conditioned on policy parameters $\theta$. The proposed controller features an ensemble of Q-function approximators $\{q_1(s, a), ..., q_n(s, a)\}$ that predict the collected Monte Carlo Q-values. The values for the update are sampled from a small experience replay buffer $\mathcal{D}_{MC}$ with recent trajectories collected by the policy. Given $n$ ensemble predictions, the controller $\mathrm{EM}$ estimates prediction uncertainty as the ensemble disagreement:
\begin{equation}
\mathrm{EM}_\omega(s|\theta) \coloneqq \Var (\{q_1(s, a), ..., q_n(s, a))\}),
\end{equation}
where $q_i(s, a) = \E_{s\sim \mathcal{D}_{MC}, a \sim \pi_\theta}[R], i\in[1..n]$. This disagreement reflects both epistemic uncertainty of controller parameters and aleatory uncertainty of the collected Monte Carlo Q-values. During optimization, controller's objective is to reduce uncertainty in a supervised fashion by minimizing square distance between the predicted and collected returns:
\begin{equation}
J_{\phi} = \sum^n_{i=1}(Q^{MC}(s, a) - q_i(s, a))^2,
\end{equation}
where $(s, a)$ is a state-action pair sampled from $\mathcal{D}_{MC}$ with the corresponding Monte Carlo Q-value $Q^{MC}(s, a)$. Taking the gradient of the controller w.r.t. action, we obtain the direction towards maximizing the uncertainty under current controller parameters $\omega$:
\begin{equation}
\nabla_a \mathrm{EM}_\omega = \nabla_a \Var(\{q_1(s, a), ..., q_n(s, a))\}) \lvert_{a\sim\pi_\theta(s).}
\end{equation}

The gradient value plays a central role in determining the exploratory action $a^\mathrm{e}$ by indicating the direction towards the least-explored regions of the environment. Figure~\ref{fig:viz_uncertainty} illustrates the surface of the uncertainty estimation value returned by $\mathrm{EM}(s)$ and the critic prediction $Q(s, \cdot)$ for the \textit{point\_mass-easy} task, where $|\mathcal{A}|=2$. The visualization is conducted after 1e5 time steps with a fixed state $s$ and a set of actions $a\in [a_{MIN}, a_{MAX}]$. The surfaces exhibit distinct highest points, resulting in different directions for $a^\mathrm{b}$ and $a^\mathrm{e}$.

We use Monte Carlo Q-value approximation as an exploratory module for two reasons. First, it is differentiable w.r.t action, allowing to obtain action gradient. Second, in contrast with model-based dynamics \citep{janner2019trust} and curiosity-based \citep{pathak2017curiosity} approaches, which require either the next state or the next action to compute the intrinsic signal, our method needs only the current state-action pair to derive a plausible action direction. As the exploratory module modifies the exploratory action, we lack access to the next transition, distinguishing our approach from methods that incorporate intrinsic signals during the off-policy update step.

\subsection{Calculating action correction}

\begin{figure}[t]
\begin{subfigure}[b]{0.22\textwidth}
    \includegraphics[width=\linewidth]{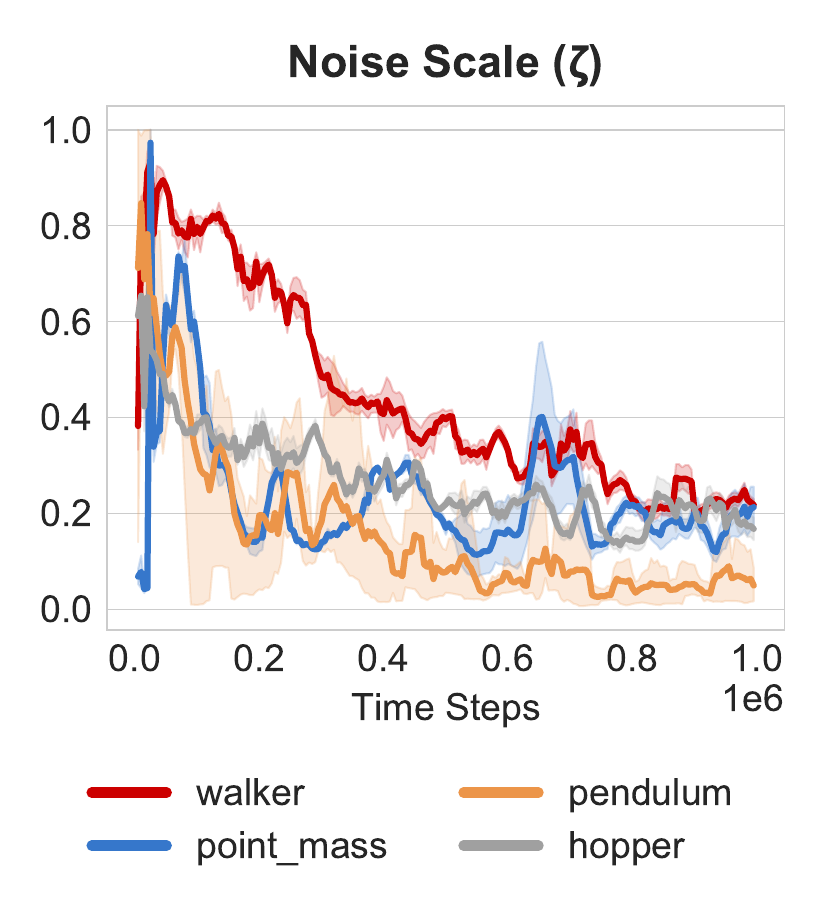}
    \caption{Dynamics of scaling coefficient $\zeta$. Values are averages across action dimensions.}
    \label{fig:scale_viz_a}
  \end{subfigure}
  \hfill 
  \begin{subfigure}[b]{0.2\textwidth}
    \includegraphics[width=\linewidth]{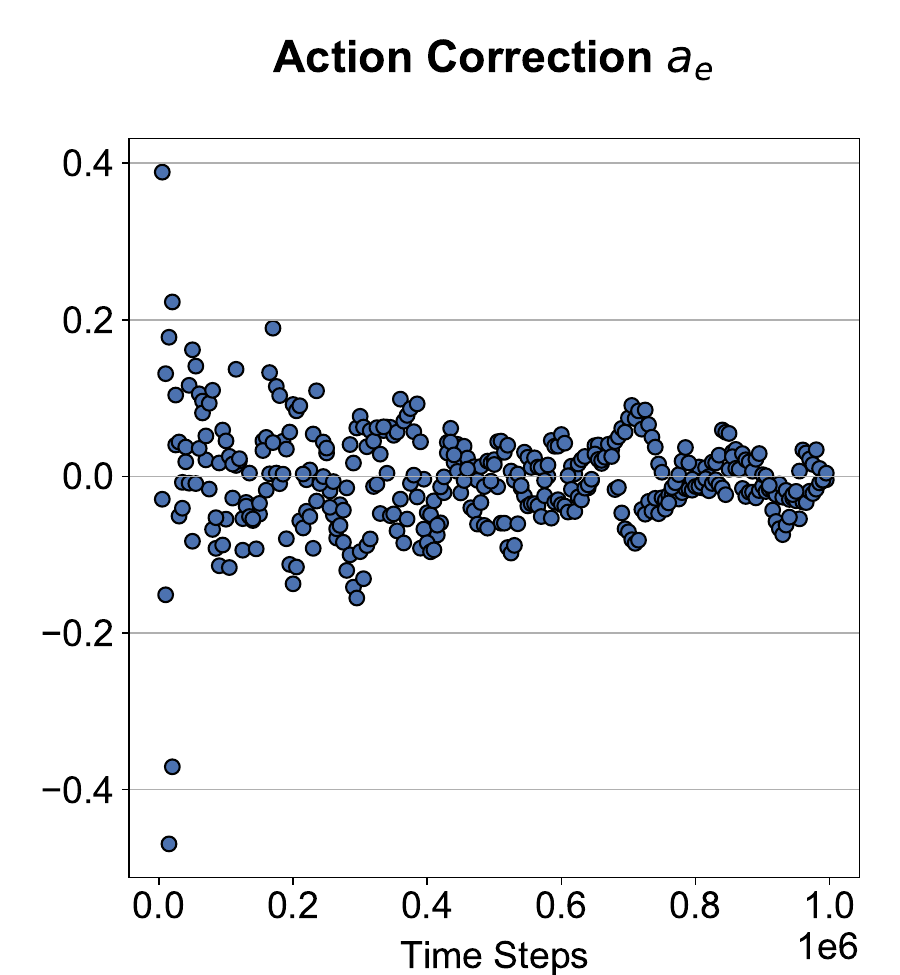}
    \caption{Dynamically calibrated exploratory action correction $a^\mathrm{e}$ during policy optimization.}
    \label{fig:scale_viz_b}
  \end{subfigure}
\caption{Visualization of exploratory action scaling.}
\end{figure}

During policy optimization the agent sequentially performs a base action $a^\mathrm{b} \sim \pi_\theta(s)$. Given the state $s$ and policy parameters $\theta$, the exploratory module provides the base action gradient towards exploration objective $\nabla_{a^\mathrm{b}}\mathrm{EM}_\omega(s|\theta)$. Here we describe the procedure of obtaining action correction $a^\mathrm{e}$ based on this gradient. 
To connect gradient value  $\nabla a^\mathrm{b}$ with action scales we perform the following scaling procedure:
\begin{equation}
a^\mathrm{e} = \frac{\nabla a^\mathrm{b}}{\lVert \nabla a^\mathrm{b} \lVert_2} \cdot \epsilon \cdot \zeta,
\end{equation}
where $\epsilon$ is denormalization constant and $\zeta$ is a scaling factor. The denormalization constant $\epsilon$ is to scale normalized gradient values to the practical magnitude of actions for a given task. It is possible to set this constant to the value that corresponds to exploration magnitude used in \citep{lillicrap2015continuous, fujimoto2018addressing}, obtaining a directional vector of a  scale required. However, with this only modification action correction vector would be fixed in its magnitude during the learning process. As a solution, we set $\epsilon$ to the magnitude of fully exploratory vector of random actions sampled uniformly:
\begin{equation}
\epsilon \coloneqq \mathbb{E}[ \lVert a \lVert_2 ], a \sim \mathcal{U}(a_{MIN}, a_{MAX})
\end{equation}
with the following multiplying by the dynamic scaling factor $\zeta$. We pose two requirements on the scaling factor: 
\begin{itemize}
    \item It should be bounded with $\zeta_i \in [0..1], i \in [1..|\mathcal{A}|]$, with $0$ corresponding to the absence of exploration and $1$ corresponding to the random action selection.
    \item It should change dynamically within the policy optimization.
\end{itemize} To satisfy both requirements we support two statistical quantities during the learning process: the running deviation of $\nabla a^\mathrm{b}$ for the last $N$ time steps $\sigma_N(\nabla a^\mathrm{b})$ and its maximum value $\max[\sigma(\nabla a^\mathrm{b})]$ for all seen deviation gradient values during the learning, resulting in:
\begin{equation}
    \zeta_i = \frac{\sigma_N(\nabla a_i^\mathrm{b})}{\max[\sigma(\nabla a_i^\mathrm{b})]}, i \in [1..|\mathcal{A}|].
\end{equation}
Large values of gradient deviation between adjacent actions from a small running window of size $N$ show uncertainty of the ensemble thus reflecting the need for more intensified exploration. The scaling factor is of dimension $|\mathcal{A}|$, facilitating keeping the individual scale for each element of a continuous action vector.

The dynamics of exploratory action $a^\mathrm{e}$ during the learning process on different tasks is depicted in Figure~\ref{fig:scale_viz_a}. In the beginning of the training, the scaling factor is close to maximum action bound enhancing aggressive exploration. During the policy and the exploratory module optimization $\zeta$ decreases leading the policy to more exploitation-inclined behavior. Notably, the dynamics of the scaling factor are not identical between environments. An example of action correction values $a^\mathrm{e}$ for the walker-run task are depicted in Figure~\ref{fig:scale_viz_b}.


\begin{figure}[t]
  \centering
  \includegraphics[width=0.9\linewidth]{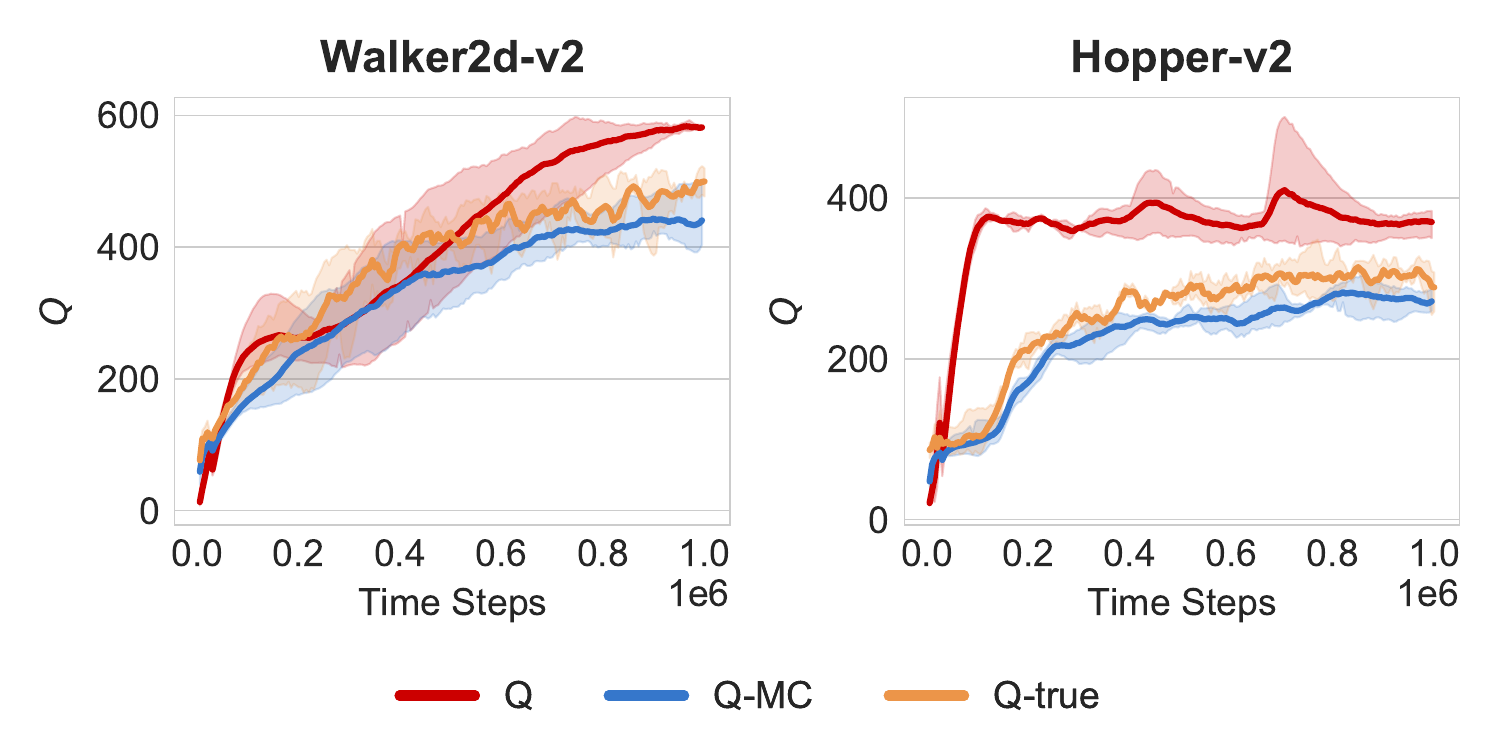}
  \caption{An illustration of Q-value overestimation. The true value of $Q$-function (Q-true) lies between overestimated critic prediction (Q) and underestimated Monte Carlo prediction (Q-MC).}
  \label{fig:overestim}
  \Description{An illustration of Q-value overestimation.}
\end{figure}

\section{MOCCO Algorithm}

Based on the presented technique of guided exploration, we propose a novel deep reinforcement learning algorithm \textbf{Mo}nte \textbf{C}arlo \textbf{C}ritic \textbf{O}ptimization (MOCCO) that incorporates an ensemble of Monte Carlo critics not only for the actor, as an exploratory module, but also for the critic to alleviate Q-value overestimation. Several works have studied Q-value overestimation in continuous control setting and have presented techniques to alleviate the issue \citep{ciosek2019better, kuznetsov2020controlling, kuznetsov2021solving}. Here, we propose to use a mean from the ensemble of Monte Carlo values as a second pessimistic Q-value estimate during critic optimization, resulting in the following critic's objective:
\begin{equation}
J_Q = (Q - Q')^2 + \beta \cdot (Q - Q^{MC})^2,
\end{equation}
where $Q^{MC}=\mu(\{q_1, ..., q_n\})$, and $\beta$ is a coefficient controlling an impact of the pessimistic estimate. 
For the given state-action pair $(s, a)$ the critic estimate $Q(s, a)$ is generally greater than the corresponding Monte Carlo estimate $Q^{MC}$, as the latter predicts estimates of the past sub-optimal policies. To illustrate this, we measure the following quantities during the learning process on tasks \textit{Walker2d-v2} and \textit{Hopper-v2} from \citep{brockman2016openai}:
\begin{itemize}
    \item TD-based bootstrapped Q-value approximation as averaged critic $Q$ prediction. Only a single critic is optimized during the TD optimization.
    \item True Q-value estimate as a discounted return of the current policy starting from given state-action pair.
    \item Monte Carlo Critic Q-value estimate as an averaged prediction of $Q^{MC}$.
\end{itemize}

\begin{figure}[t]
  \centering
  \includegraphics[width=0.6\linewidth]{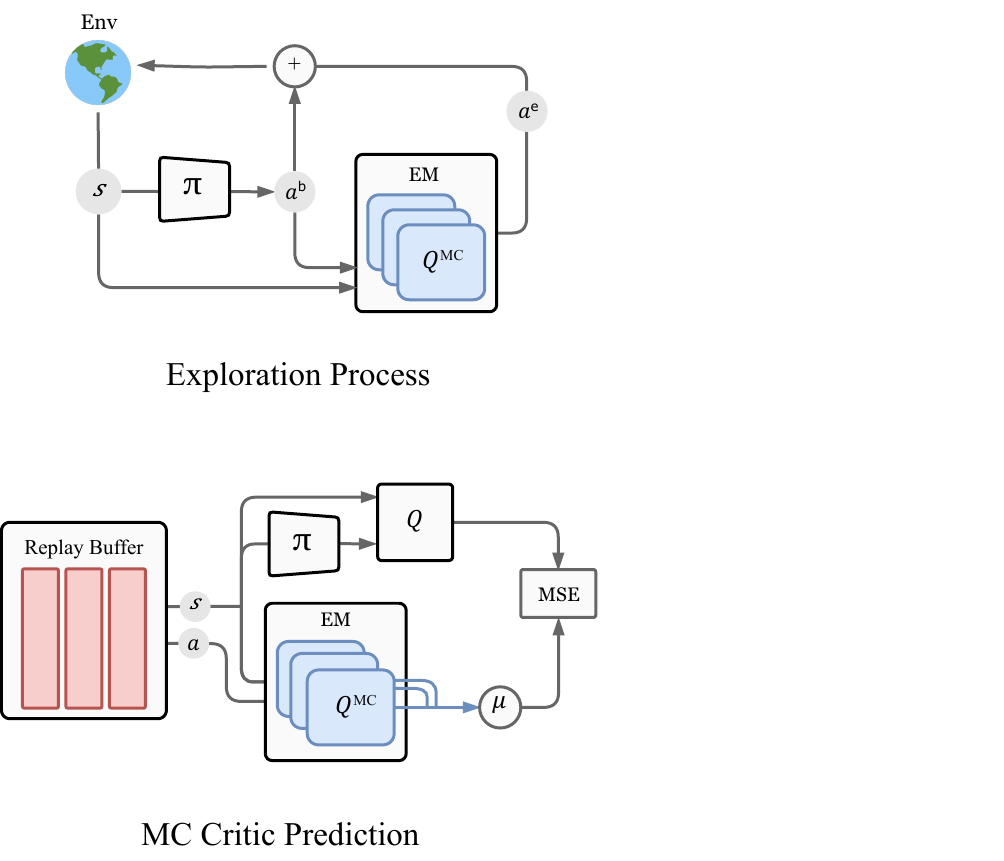}
  \caption{MOCCO: A role of the exploratory module (EM) during exploration (top) and critic optimization (bottom).}
  \label{fig:arch}
  \Description{MOCCO Algorithm.}
\end{figure}

\begin{algorithm}[t]
\caption{MOCCO}\label{alg:mocco}
\begin{algorithmic}[1]
   \STATE Initialize actor \(\pi_{\theta^\pi}\), critic \(Q_{\theta^Q}\), target critic \(Q^{'}_{\theta^{Q^{'}}}\)
   \STATE Initialize exploratory module \(\mathrm{EM}_\omega=\{q_1,...,q_k\}\)
   \STATE Initialize replay buffer \(\mathcal{B}\), episodic buffer \(\mathcal{B}_E\)
   \STATE Initialize smaller buffer \(\mathcal{B}^{MC}\) for the most recent steps
   \FOR{t=1 to T}
   \STATE Select action with guided noise $a = a^\mathrm{b} + a^\mathrm{e}$,
   \STATE $a^\mathrm{b} \sim \pi_{\theta^\pi}(s), a^\mathrm{e} \sim \mathrm{EM}_{\omega}$
   \STATE Receive next state \(s^{'}\), reward \(r\), terminal signal \(d\)
   \IF{\(d\)}
   \FOR{i=1 to \(|\mathcal{B}_E|\)}
   \STATE Calculate discounted episodic return \(R_i\)
   \STATE Store \((s_i, a_i, r_i, s^{'}_i, d_i)\) in \(\mathcal{B}\)
   \STATE Store \((s_i, a_i, R_i)\) in \(\mathcal{B}^{MC}\)
   \ENDFOR
   \STATE Free $\mathcal{B}_E$
   \ENDIF
   \STATE Sample mini-batch from \(\mathcal{B}\): \([s_b, a_b, r_b, s_b]\) and
   \STATE update critic \(Q\): $ J_{Q} = (Q - Q')^2 + \beta (Q - Q^{MC})^2 $
   \STATE Sample mini-batch from \(\mathcal{B}^{MC}\): \([s_b, a_b, R_b]\) and
   \STATE update $\mathrm{EM}_\omega$: $J_{\mathrm{EM}}=\sum_i^k (q_i - R_b)^2$
   \IF{t mod 2}
   \STATE Update policy \(\pi_\theta^{\pi}\)
   \STATE Update target critic \(\theta^Q{^{'}} \leftarrow \tau \theta^Q + (1-\tau) \theta^{Q^{'}}\)
   \ENDIF
   \ENDFOR
   \STATE \textbf{return} \(\theta^\pi\)
\end{algorithmic}
\end{algorithm}

All three quantities are averages across the batch size of 256 that are collected each 5e3 time steps and reported from multiple seeds. Figure~\ref{fig:overestim} shows the dynamics of predictions on 1M time steps. Monte Carlo prediction (Q-MC) has lower value than both TD-based (Q) and true Q-estimate (Q-true) values, while TD-based prediction is generally higher than the true estimate.

In reinforcement learning, Monte Carlo estimates have high variance and low bias, whereas one-step TD methods have less variance but can be biased. Here, we combine both methods during Q-function optimization. Empirically we show that the true Q-value estimate lies between $Q$ and $Q^{MC}$, therefore it is beneficial to use both estimates to balance between over- and underestimation.

The role of the exploratory module (EM) during the exploration and critic optimization is schematically depicted in Figure~\ref{fig:arch}. During action selection, EM provides directed exploration i.e. it is applied on-policy. During off-policy update steps, EM is used to provide a second estimate for current Q-function optimization.

Practically, the MOCCO algorithm is based on TD3 with the following differences: (1) it uses guided exploration during action selection; (2) it does not use second TD critic; (3) the mean of MC-critics ensemble is used during the critic optimization. Full pseudo-code is presented in Algorithm~\ref{alg:mocco}.


\section{Experiments}

First, we demonstrate that the proposed method improves conventional random noise-based exploration using TD3 algorithm. Next, we show experimental results for MOCCO. We perform an ablation study to identify the impact of each proposed algorithmic component on final success and demonstrate the comparative evaluation results.

\subsection{Guided Exploration for Off-Policy Deterministic Algorithms}

In this experiment, we show that guided exploration improves over the traditional Gaussian-based exploration method. To do so, we use TD3 as a baseline algorithm and vary the type of applied exploratory noise during the action selection process. Table~\ref{tab:exploration} shows the results for 8 control tasks from DMControl Suite. The studied exploration methods are as following: actions are sampled without noise (NO EXPL), with Normal Gaussian noise (NORM), temporally correlated noise drawn from an Ornstein-Uhlenbeck process (OU), curiosity-based exploration with RND method from \citep{burda2018exploration}, and with proposed guided exploration method (GE). 

For normal Gaussian and Ornstein-Uhlenbeck noise, we use parameters reported in \citep{haarnoja2018soft, lillicrap2015continuous} correspondingly. For the RND method, we add an intrinsic reward term to the original reward which may facilitate exploration greatly in the case of sparse signal environments. For the target and predictor networks, we use the fully connected MLP of dimensions (128, 64) with the same learning rate used for the actor and the critic. Following practical details from \citep{burda2018exploration}, we perform state normalization during the calculation of the target-predictor error based on the last 1000 environment transitions. We run each algorithm 10 times with different random seeds. Within each run, the reported value is the mean of 10 evaluations of the same policy from different environment initialization. Each evaluation value is the mean of the last 10 episodes for stability purposes. 

Results demonstrate higher rewards of guided exploration over the conventional approaches for all environments. The absence of exploration provides better results than the normal noise in half of the environments presented, which empirically proves our initial suggestion that normal noise often can hurt the performance of an agent. The curiosity-based exploration approach also provides better results than the normal noise in half of the environments.

\begin{table}[t]
	\caption{A comparison of exploration approaches. Average episodic score from 10 trials. Each trial is a mean of the last 10 episodes.}
	\label{tab:exploration}
	\begin{tabular}{rlllll}\toprule
		\textit{Task} & \textit{NO EXPL} & \textit{NORM}  & \textit{OU} & \textit{RND} & \textit{GE} \\ \midrule
		acrobot       & 0.65 & 0.59 & 1.67 & 0.67 & \bfseries{150.46} \\
		point\_mass   & 726.80 & 709.90 & 605.09 & 	743.36 & \bfseries{785.25} \\
		pendulum      & 247.411 & 359.94 & 215.13 & 241.13 & \bfseries{792.33} \\
		walker-walk   & 921.80 & 940.80 & 953.13 & 881.04 & \bfseries{955.977} \\
	    walker-run    & 644.32 & 596.71 & 622.06 & 589.50 & \bfseries{677.73} \\ 
        hopper-stand  & 56.40 & 34.86 & 58.67 & 51.91 & \bfseries{602.14} \\ 
        hopper-hop    & 18.30 & 27.18 & 16.31 & 14.14 & \bfseries{167.12} \\ 
        human-walk    & 3.32 & 79.16 & 30.02 & 129.83 & \bfseries{329.25} \\ \bottomrule
	\end{tabular}
\end{table}

\subsection{MOCCO Evaluation}

We demonstrate comparative results of MOCCO algorithm on 14 control problems from DMControl benchmark for the following domains: acrobot, fish, hopper, humanoid, pendulum, point\_mass, swimmer, and walker. We compare the proposed approach with off-policy algorithms DDPG, TD3, SAC \citep{pmlr-v80-haarnoja18b}. We run 1M environment time steps on each environment, collecting episode scores for every 2e3 steps. Each episode score is an average of 10 runs of a policy without exploration from different environment seeds. Each algorithm is run 10 times with different seeds with reported standard deviation.

We use the official implementation of TD3 as a baseline. For DDPG, we use the implementation from TD3, denoted as "DDPG". We extend the TD3 code to entropy optimization for SAC deriving. For TD3-RND, we implement target-predicted module from \citep{burda2018exploration} and combine extrinsic reward with intrinsic signal.

\begin{figure*}[t]
  \centering
  \includegraphics[width=0.9\linewidth]{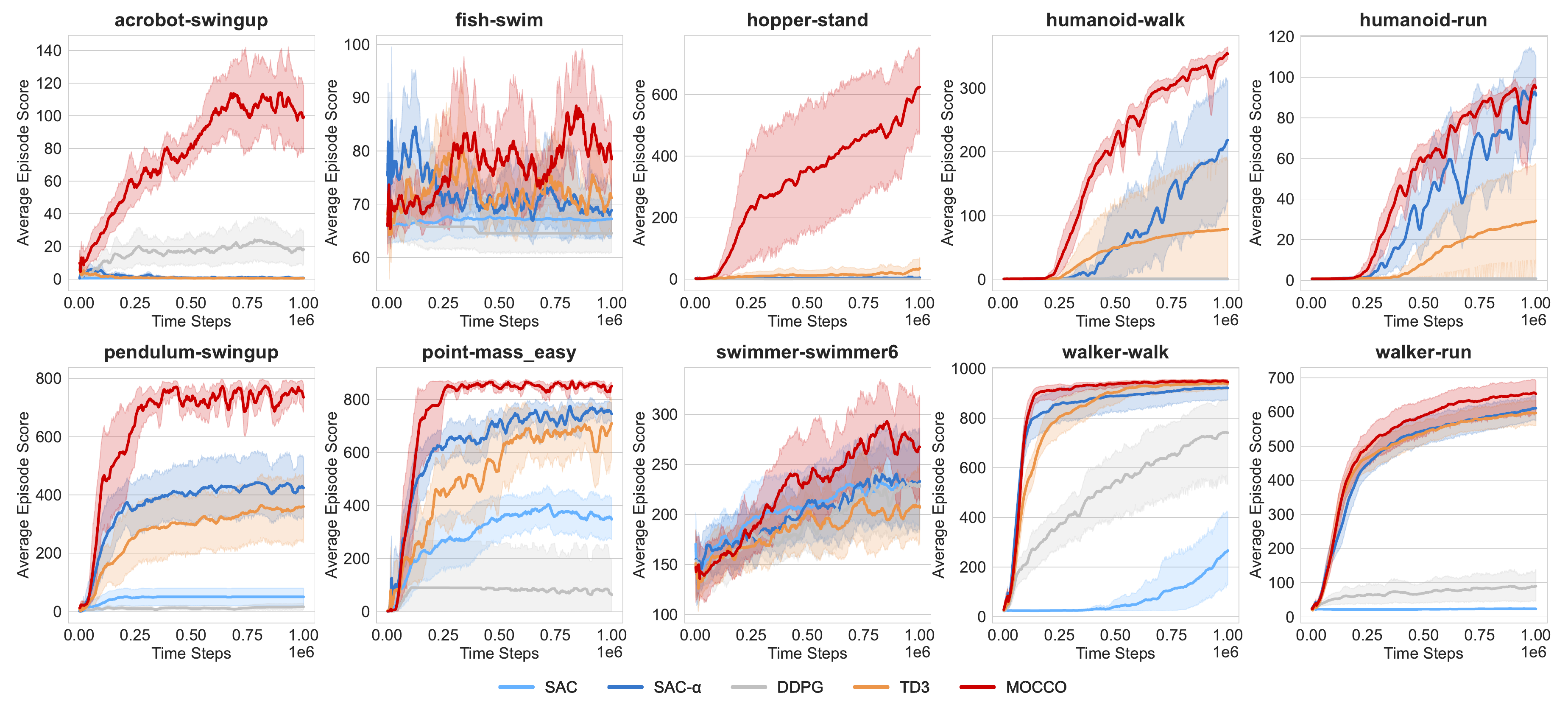}
  \caption{Learning curves of evaluation algorithms.}
  \label{fig:eval}
  \Description{earning curves of evaluation algorithms.}
\end{figure*}

\begin{table*}[t]
\caption{Evaluation results as the average of the last 10 episodes. $\pm$ denotes standard deviation across seeds.}
\label{table_eval}
\vskip 0.15in
\begin{center}
\begin{small}
\begin{sc}
\begin{tabular}{lcccccr}
\toprule
Env & SAC & SAC-$\alpha$  & DDPG & TD3 & TD3-RND & MOCCO \\
\midrule
acrobot-swingup               & 0.666     & 0.757     & 18.3405   & 0.598      & 0.671 & \bfseries{99.047 $\pm$ 32.012}  \\
acrobot-swingup\_sparse       & 0.000     & 0.000     & 0.373     & 0 .000     & 0.000 & \bfseries{7.805 $\pm$ 4.441}  \\
fish-swim                     & 67.278    & 69.149    & 64.550    & 71.635     & 74.237 & \bfseries{78.137 $\pm$ 11.678}  \\
hopper-stand                  & 3.937     & 5.461     & 0.801     & 34.861     & 51.915 & \bfseries{624.478 $\pm$ 226.007}  \\
hopper-hop                    & 0.378     & 29.334    & 0.097     & 27.182     & 14.149 & \bfseries{166.969 $\pm$ 105.5}  \\
humanoid-walk                 & 0.859     & 218.491   & 1.232     & 79.162     & 129.837 & \bfseries{353.124 $\pm$ 15.761}  \\
humanoid-run                  & 0.809     & 90.760    & 0.783     & 29.322     & 13.620 & \bfseries{94.783 $\pm$ 6.838}  \\
pendulum-swingup              & 50.000    & 424.045   & 15.851    & 359.941    & 241.130 & \bfseries{735.768 $\pm$ 73.214}  \\
point-mass\_easy              & 345.787   & 742.655   & 61.386    & 709.902    & 743.365 & \bfseries{852.272 $\pm$ 26.531}  \\
finger-turn\_easy              & 478.109   & 597.356   & 192.284   & 648.087    & 530.673 & \bfseries{674.340 $\pm$ 191.697 } \\
swimmer-swimmer6              & 228.513   & 233.111   & 226.501   & 206.300    & 217.624 & \bfseries{268.100 $\pm$ 80.347}  \\
swimmer-swimmer15             & 98.288    & 103.964   & 103.778   & 91.867     & 92.019 & \bfseries{107.265 $\pm$ 39.124}  \\
walker-walk                   & 266.182   & 922.416   & 744.633   & 940.802    & 918.704 & \bfseries{946.637 $\pm$ 19.110}  \\
walker-run                    & 23.585    & 610.246   & 68.455    & 596.718    & 589.503 & \bfseries{651.953 $\pm$ 83.979}  \\
\bottomrule
\end{tabular}
\end{sc}
\end{small}
\end{center}
\vskip -0.1in
\end{table*}

We use Adam optimizer \citep{kingma2014adam} with the learning rate 3e-4 for all algorithms. For a fair comparison, we use the same network parameters from the \citep{fujimoto2018addressing}. All networks have 2 hidden layers of size 256 and ReLU non-linearity \citep{glorot2011deep}. For SAC we run two versions, first with fixed entropy coefficient equal to 0.2 (proposed in \citep{SpinningUp2018}) and auto tunable entropy (SAC-$\alpha$) from \citep{haarnoja2018soft}. We use mini-batch size 256 for all algorithms. 

For MOCCO, we set the size of $\mathcal{D}_{MC}$ to 1e5 transitions, and the size of buffer $N$ for calculating deviations used in the scaling factor $\zeta$ to 10. We use 3 critics in the ensemble as our experiments have not provided benefits from an increased number of critics. The $\beta$ coefficient controlling an impact of MC-critic is set to either 0.1 or 0.01 depending from the best performance.

Table~\ref{table_eval} reports the average of episodic scores for the last 10 episodes. The learning curves for each domain are presented in Figure~\ref{fig:eval}. MOCCO outperforms other approaches, sometimes with significant margins, e.g. for acrobot, pendulum, and hopper domains.

\subsection{MOCCO: Ablation Study}

The aim of this experiment is to identify the impact of each component on the overall success of the presented algorithm. We study the contribution of guided exploration as a substitution of the Gaussian noise (TD3 + GE), the contribution of the proposed critic objective featuring Monte Carlo critic estimate (TD3 + QMC), and of both features combined (MOCCO).

\begin{figure}[t]
  \centering
  \includegraphics[width=0.8\linewidth]{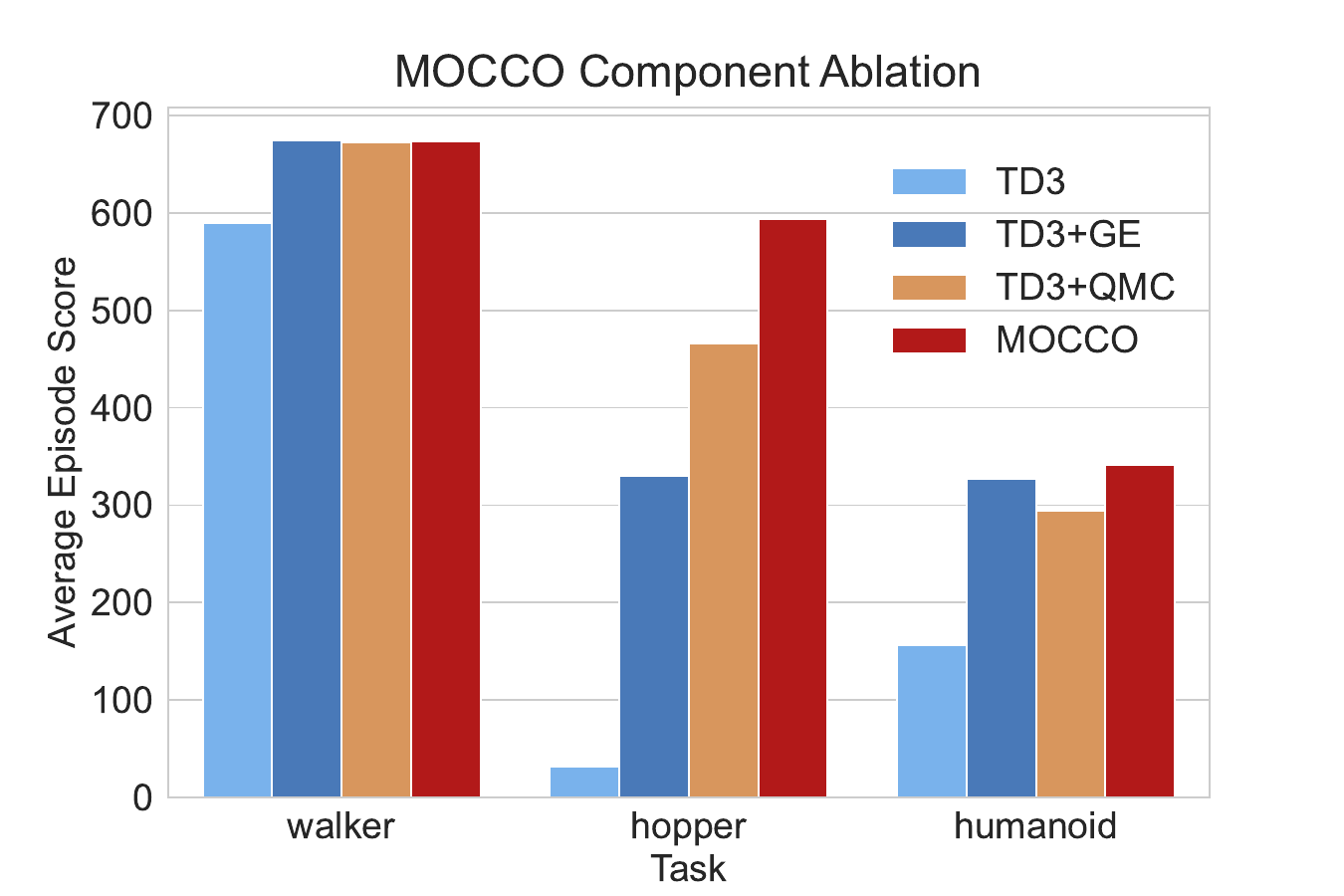}
  \caption{Contribution of different components of MOCCO algorithm.}
  \label{fig:contrib}
  \Description{Contribution of different components of MOCCO algorithm.}
\end{figure}

Figure~\ref{fig:contrib} shows results for walker-run, hopper-stand, and humanoid-walk on 5 random seeds. Results show that each component improves the results of the baseline algorithm on all tasks and a combination of features provides better performance than their sole contributions on 2 out of 3 tasks.

\begin{figure}[t]
\centering
\begin{subfigure}{\columnwidth}
\centering
\includegraphics[width=\columnwidth]{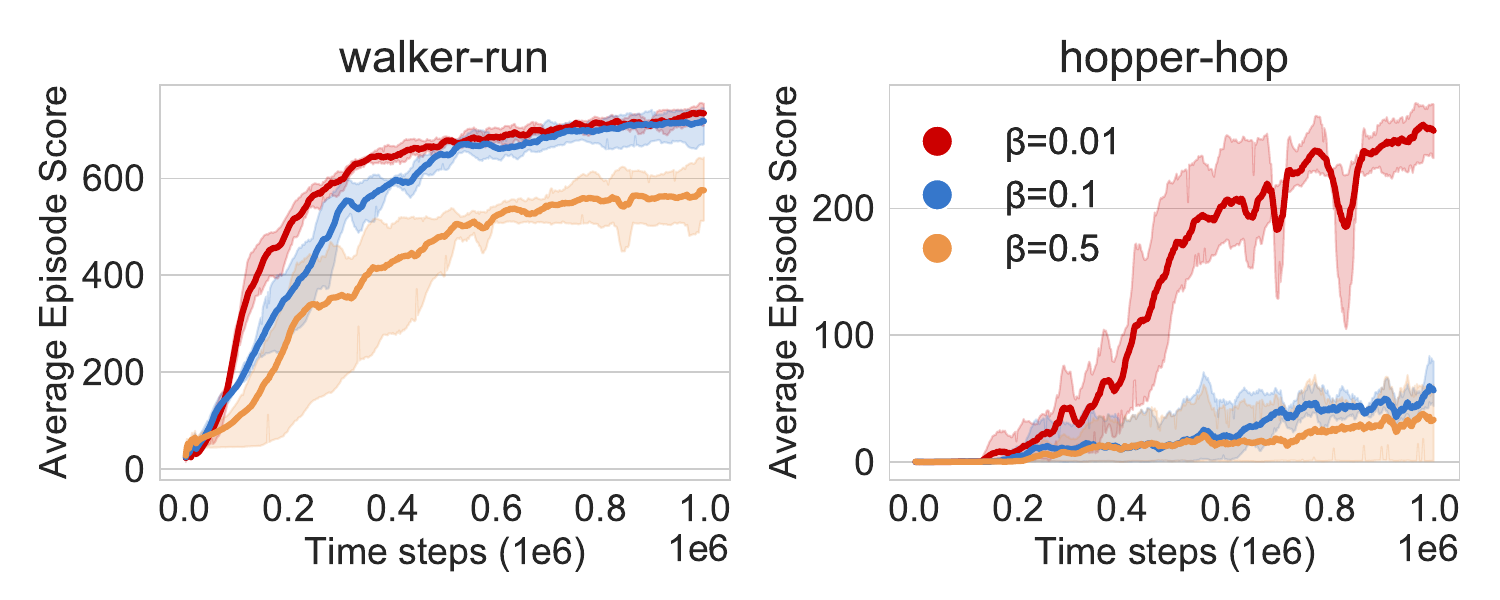}
\caption{$\beta$ hyperparameter.}
\end{subfigure}
\hfill
\begin{subfigure}{\columnwidth}
\centering
\includegraphics[width=\columnwidth]{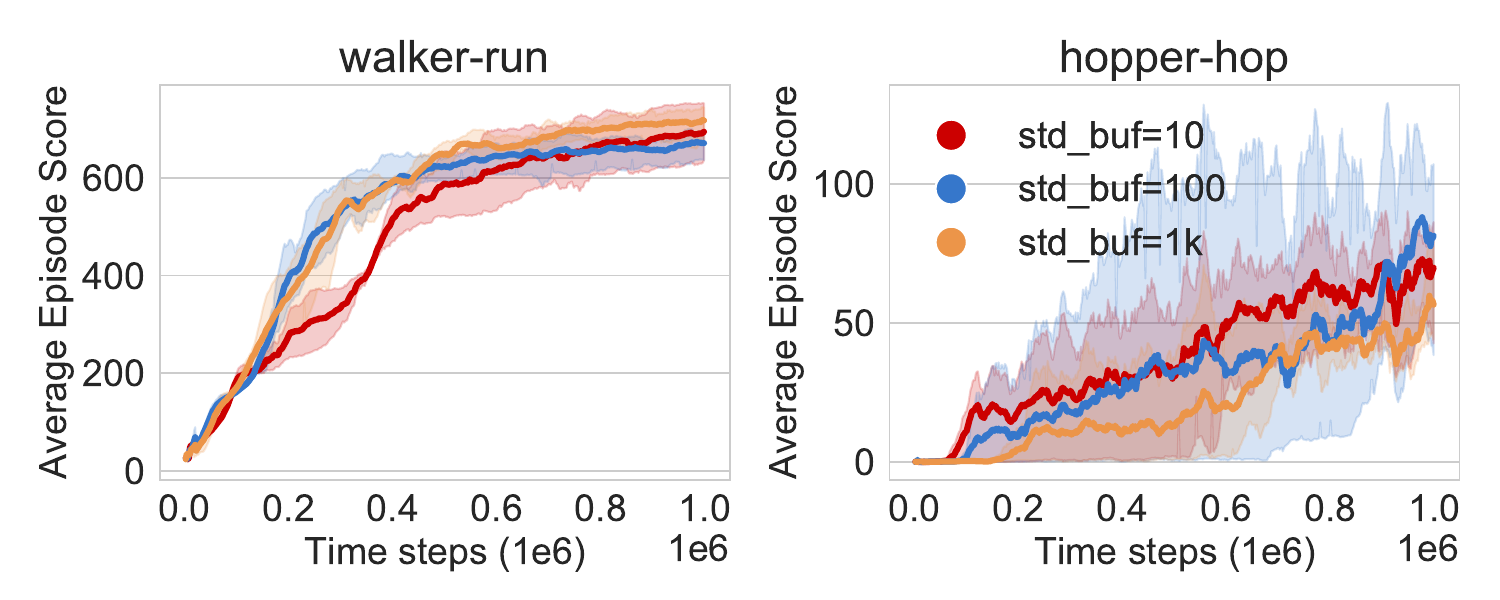}
\caption{The size $N$ of the running window for calculating $\sigma_N(\nabla a)$.}
\end{subfigure}
\hfill
\begin{subfigure}{\columnwidth}
\centering
\includegraphics[width=\columnwidth]{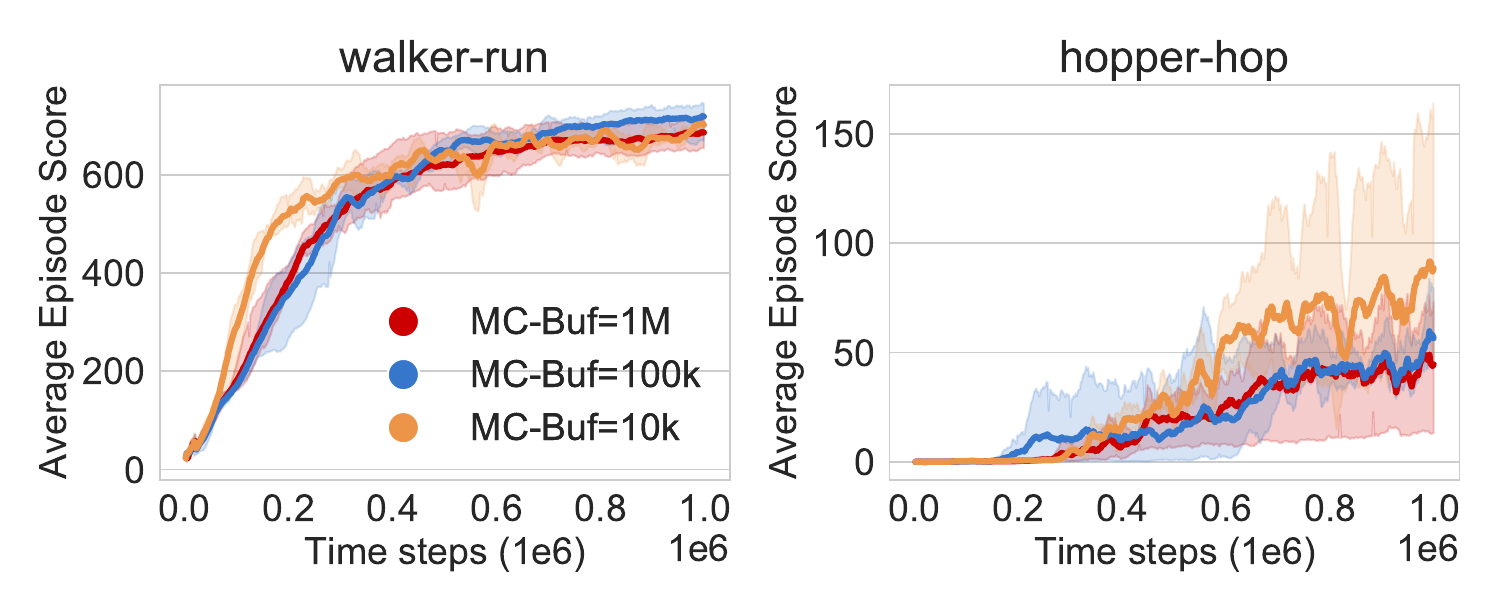}
\caption{The size of $\mathcal{D}_{MC}$.}
\end{subfigure}
\caption{The impact of different hyperparameter values.}
\label{mocco_hyps}
\end{figure}

Next, we vary hyperparameters introduced by MOCCO to identify algorithm robustness on two tasks: walker-walk and hopper-hop. We study different values of the coefficient controlling MC critic impact ($\beta$), the size $N$ of the buffer used for calculating the running deviation of action gradient $\sigma_N(\nabla a)$, and the size of reduced replay buffer $\mathcal{B}^{MC}$ on which the ensemble of MC critics is trained. Figure~\ref{mocco_hyps} shows that MOCCO is robust to the hyperparameter change, except  the case of varying $\beta$ for hopper-hop.


\section{Related Work}

The problem of efficient exploration has a long-standing history in reinforcement learning. Early approaches suggest incorporating random exploratory strategies \citep{moore1990efficient,sutton1990integrated,barto1991real,williams1992simple} which have theoretical grounds but are not always scalable to complex environments with high-dimensional inputs. The broad class of directed exploration algorithms, first proposed in \citep{thrun1992efficient}, studies an approach of guiding a policy towards certain regions with specific knowledge of the learning process.

Numerous works on directed exploration study intrinsic reward, where the learning process relies on some intrinsic signal. Examples of exploration based on intrinsic reward include algorithms based on information gain \citep{little2013learning,mobin2014information}, curiosity \citep{deci1981characteristics,pathak2017curiosity}, and uncertainty-minimization \citep{schmidhuber1991curious,houthooft2016vime}. \citep{badia2019never} proposes to use episodic memory as a source of the intrinsic signal and optimize multiple policies, some of which are guided by exploratory goals and others are fully exploitative. In contrast with many intrinsic-reward approaches, our method does not modify the reward with additional dense signal nor modifies the policy, but rather optimizes the intrinsic objective independently only to correct exploratory action.

The work of \citep{pathak2019self} shows that exploration can be key in achieving high sample efficiency required for real-world setup. The authors use an ensemble of forward dynamics models that produces disagreement as an intrinsic reward. Incorporation of such a disagreement signal results in an effective exploration, allowing robots to solve object interaction tasks from scratch in real-time.

Apart from the directed exploration based on intrinsic objective, several works have proposed methods with a different perspective. \citep{ciosek2019better} addresses the notion of \emph{pessimistic underexploration} that comes from optimizing a lower bound in critic update. By approximating the upper bound of state-action function and using it during exploration the proposed method avoids underexploration and achieves higher sample efficiency. Some works study the approach of exploration in policy parameter space rather than in action space \citep{van2017generalized,fortunato2018noisy,pmlr-v139-zhang21t}. Finally, a number of works study exploration under goal-reaching angle, where the agent explores environment by learning to achieve seen \citep{florensa2018automatic}, generated \citep{nair2018visual,sukhbaatar2018intrinsic}, or specifically unseen \citep{mendonca2021discovering} state regions.

A comprehensive review of exploration in reinforcement learning can be found in \citep{amin2021survey}. Following the suggested taxonomy, the method presented in this paper falls into the category of "reward-free intrinsically-motivated" exploration i.e. not using an extrinsic reward for exploratory purpose, but relying on intrinsic signal to explore novel world regions.

\section{Discussion}

We have introduced a method for directed exploration that enhances off-policy deterministic algorithms by incorporating exploratory action correction into the base policy action. Utilizing a differential controller conditioned on policy parameters, we guide exploration towards regions of interest and dynamically adjust the magnitude of exploratory correction throughout the learning process. The proposed exploration method avoids the need for extensive hyperparameter tuning, as it relies on a learnable intrinsic signal reflecting environmental uncertainty. Empirical results demonstrate the effectiveness of our approach across various continuous control tasks, showcasing superior rewards compared to methods relying on random noise. Additionally, we present MOCCO, a novel deep reinforcement learning algorithm based on our proposed exploration method. Experimental results indicate that MOCCO outperforms modern off-policy actor-critic methods across a set of DMControl suite tasks.

For instance, this controller could be formulated as a form of forward or inverse model dynamics, thus redirecting attention towards model-based reinforcement learning. Another promising avenue is to conceptualize the exploratory module as a differential memory module, aligning the direction of the external agent's memory with efficient exploration. We believe that the proposed exploration technique will contribute to addressing challenges that demand thoughtful dynamic exploration.



\begin{acks}
The authors would like to thank Dmitry Akimov for insightful discussions. We thank Andrey Filchenkov for fruitful feedback during the early-stage development of the method.
\end{acks}



\bibliographystyle{ACM-Reference-Format} 
\bibliography{sample}


\begin{thebibliography}{47}


\ifx \showCODEN    \undefined \def \showCODEN     #1{\unskip}     \fi
\ifx \showDOI      \undefined \def \showDOI       #1{#1}\fi
\ifx \showISBNx    \undefined \def \showISBNx     #1{\unskip}     \fi
\ifx \showISBNxiii \undefined \def \showISBNxiii  #1{\unskip}     \fi
\ifx \showISSN     \undefined \def \showISSN      #1{\unskip}     \fi
\ifx \showLCCN     \undefined \def \showLCCN      #1{\unskip}     \fi
\ifx \shownote     \undefined \def \shownote      #1{#1}          \fi
\ifx \showarticletitle \undefined \def \showarticletitle #1{#1}   \fi
\ifx \showURL      \undefined \def \showURL       {\relax}        \fi
\providecommand\bibfield[2]{#2}
\providecommand\bibinfo[2]{#2}
\providecommand\natexlab[1]{#1}
\providecommand\showeprint[2][]{arXiv:#2}

\bibitem[\protect\citeauthoryear{Achiam}{Achiam}{2018}]%
        {SpinningUp2018}
\bibfield{author}{\bibinfo{person}{Joshua Achiam}.} \bibinfo{year}{2018}\natexlab{}.
\newblock \showarticletitle{{Spinning Up in Deep Reinforcement Learning}}.
\newblock  (\bibinfo{year}{2018}).
\newblock


\bibitem[\protect\citeauthoryear{Amin, Gomrokchi, Satija, van Hoof, and Precup}{Amin et~al\mbox{.}}{2021}]%
        {amin2021survey}
\bibfield{author}{\bibinfo{person}{Susan Amin}, \bibinfo{person}{Maziar Gomrokchi}, \bibinfo{person}{Harsh Satija}, \bibinfo{person}{Herke van Hoof}, {and} \bibinfo{person}{Doina Precup}.} \bibinfo{year}{2021}\natexlab{}.
\newblock \showarticletitle{A Survey of Exploration Methods in Reinforcement Learning}.
\newblock \bibinfo{journal}{\emph{arXiv preprint arXiv:2109.00157}} (\bibinfo{year}{2021}).
\newblock


\bibitem[\protect\citeauthoryear{Badia, Sprechmann, Vitvitskyi, Guo, Piot, Kapturowski, Tieleman, Arjovsky, Pritzel, Bolt, et~al\mbox{.}}{Badia et~al\mbox{.}}{2019}]%
        {badia2019never}
\bibfield{author}{\bibinfo{person}{Adri{\`a}~Puigdom{\`e}nech Badia}, \bibinfo{person}{Pablo Sprechmann}, \bibinfo{person}{Alex Vitvitskyi}, \bibinfo{person}{Daniel Guo}, \bibinfo{person}{Bilal Piot}, \bibinfo{person}{Steven Kapturowski}, \bibinfo{person}{Olivier Tieleman}, \bibinfo{person}{Martin Arjovsky}, \bibinfo{person}{Alexander Pritzel}, \bibinfo{person}{Andrew Bolt}, {et~al\mbox{.}}} \bibinfo{year}{2019}\natexlab{}.
\newblock \showarticletitle{Never Give Up: Learning Directed Exploration Strategies}. In \bibinfo{booktitle}{\emph{International Conference on Learning Representations}}.
\newblock


\bibitem[\protect\citeauthoryear{Barto, Bradtke, and Singh}{Barto et~al\mbox{.}}{1991}]%
        {barto1991real}
\bibfield{author}{\bibinfo{person}{Andrew~Gehret Barto}, \bibinfo{person}{Steven~J Bradtke}, {and} \bibinfo{person}{Satinder~P Singh}.} \bibinfo{year}{1991}\natexlab{}.
\newblock \bibinfo{booktitle}{\emph{Real-time learning and control using asynchronous dynamic programming}}.
\newblock \bibinfo{publisher}{University of Massachusetts at Amherst, Department of Computer and~…}.
\newblock


\bibitem[\protect\citeauthoryear{Berlyne}{Berlyne}{1966}]%
        {berlyne1966curiosity}
\bibfield{author}{\bibinfo{person}{Daniel~E Berlyne}.} \bibinfo{year}{1966}\natexlab{}.
\newblock \showarticletitle{Curiosity and exploration}.
\newblock \bibinfo{journal}{\emph{Science}} \bibinfo{volume}{153}, \bibinfo{number}{3731} (\bibinfo{year}{1966}), \bibinfo{pages}{25--33}.
\newblock


\bibitem[\protect\citeauthoryear{Brockman, Cheung, Pettersson, Schneider, Schulman, Tang, and Zaremba}{Brockman et~al\mbox{.}}{2016}]%
        {brockman2016openai}
\bibfield{author}{\bibinfo{person}{Greg Brockman}, \bibinfo{person}{Vicki Cheung}, \bibinfo{person}{Ludwig Pettersson}, \bibinfo{person}{Jonas Schneider}, \bibinfo{person}{John Schulman}, \bibinfo{person}{Jie Tang}, {and} \bibinfo{person}{Wojciech Zaremba}.} \bibinfo{year}{2016}\natexlab{}.
\newblock \showarticletitle{Openai gym}.
\newblock \bibinfo{journal}{\emph{arXiv preprint arXiv:1606.01540}} (\bibinfo{year}{2016}).
\newblock


\bibitem[\protect\citeauthoryear{Burda, Edwards, Storkey, and Klimov}{Burda et~al\mbox{.}}{2018}]%
        {burda2018exploration}
\bibfield{author}{\bibinfo{person}{Yuri Burda}, \bibinfo{person}{Harrison Edwards}, \bibinfo{person}{Amos Storkey}, {and} \bibinfo{person}{Oleg Klimov}.} \bibinfo{year}{2018}\natexlab{}.
\newblock \showarticletitle{Exploration by random network distillation}.
\newblock \bibinfo{journal}{\emph{arXiv preprint arXiv:1810.12894}} (\bibinfo{year}{2018}).
\newblock


\bibitem[\protect\citeauthoryear{Ciosek, Vuong, Loftin, and Hofmann}{Ciosek et~al\mbox{.}}{2019}]%
        {ciosek2019better}
\bibfield{author}{\bibinfo{person}{Kamil Ciosek}, \bibinfo{person}{Quan Vuong}, \bibinfo{person}{Robert Loftin}, {and} \bibinfo{person}{Katja Hofmann}.} \bibinfo{year}{2019}\natexlab{}.
\newblock \showarticletitle{Better exploration with optimistic actor-critic}. In \bibinfo{booktitle}{\emph{Proceedings of the 33rd International Conference on Neural Information Processing Systems}}. \bibinfo{pages}{1787--1798}.
\newblock


\bibitem[\protect\citeauthoryear{Deci, Nezlek, and Sheinman}{Deci et~al\mbox{.}}{1981}]%
        {deci1981characteristics}
\bibfield{author}{\bibinfo{person}{Edward~L Deci}, \bibinfo{person}{John Nezlek}, {and} \bibinfo{person}{Louise Sheinman}.} \bibinfo{year}{1981}\natexlab{}.
\newblock \showarticletitle{Characteristics of the rewarder and intrinsic motivation of the rewardee.}
\newblock \bibinfo{journal}{\emph{Journal of personality and social psychology}} \bibinfo{volume}{40}, \bibinfo{number}{1} (\bibinfo{year}{1981}), \bibinfo{pages}{1}.
\newblock


\bibitem[\protect\citeauthoryear{Florensa, Held, Geng, and Abbeel}{Florensa et~al\mbox{.}}{2018}]%
        {florensa2018automatic}
\bibfield{author}{\bibinfo{person}{Carlos Florensa}, \bibinfo{person}{David Held}, \bibinfo{person}{Xinyang Geng}, {and} \bibinfo{person}{Pieter Abbeel}.} \bibinfo{year}{2018}\natexlab{}.
\newblock \showarticletitle{Automatic goal generation for reinforcement learning agents}. In \bibinfo{booktitle}{\emph{International conference on machine learning}}. PMLR, \bibinfo{pages}{1515--1528}.
\newblock


\bibitem[\protect\citeauthoryear{Fortunato, Azar, Piot, Menick, Hessel, Osband, Graves, Mnih, Munos, Hassabis, et~al\mbox{.}}{Fortunato et~al\mbox{.}}{2018}]%
        {fortunato2018noisy}
\bibfield{author}{\bibinfo{person}{Meire Fortunato}, \bibinfo{person}{Mohammad~Gheshlaghi Azar}, \bibinfo{person}{Bilal Piot}, \bibinfo{person}{Jacob Menick}, \bibinfo{person}{Matteo Hessel}, \bibinfo{person}{Ian Osband}, \bibinfo{person}{Alex Graves}, \bibinfo{person}{Volodymyr Mnih}, \bibinfo{person}{Remi Munos}, \bibinfo{person}{Demis Hassabis}, {et~al\mbox{.}}} \bibinfo{year}{2018}\natexlab{}.
\newblock \showarticletitle{Noisy Networks For Exploration}. In \bibinfo{booktitle}{\emph{International Conference on Learning Representations}}.
\newblock


\bibitem[\protect\citeauthoryear{Fujimoto, Hoof, and Meger}{Fujimoto et~al\mbox{.}}{2018}]%
        {fujimoto2018addressing}
\bibfield{author}{\bibinfo{person}{Scott Fujimoto}, \bibinfo{person}{Herke Hoof}, {and} \bibinfo{person}{David Meger}.} \bibinfo{year}{2018}\natexlab{}.
\newblock \showarticletitle{Addressing function approximation error in actor-critic methods}. In \bibinfo{booktitle}{\emph{International Conference on Machine Learning}}. PMLR, \bibinfo{pages}{1587--1596}.
\newblock


\bibitem[\protect\citeauthoryear{Garc{\i}a and Fern{\'a}ndez}{Garc{\i}a and Fern{\'a}ndez}{2015}]%
        {garcia2015comprehensive}
\bibfield{author}{\bibinfo{person}{Javier Garc{\i}a} {and} \bibinfo{person}{Fernando Fern{\'a}ndez}.} \bibinfo{year}{2015}\natexlab{}.
\newblock \showarticletitle{A comprehensive survey on safe reinforcement learning}.
\newblock \bibinfo{journal}{\emph{Journal of Machine Learning Research}} \bibinfo{volume}{16}, \bibinfo{number}{1} (\bibinfo{year}{2015}), \bibinfo{pages}{1437--1480}.
\newblock


\bibitem[\protect\citeauthoryear{Gershman}{Gershman}{2019}]%
        {gershman2019uncertainty}
\bibfield{author}{\bibinfo{person}{Samuel~J Gershman}.} \bibinfo{year}{2019}\natexlab{}.
\newblock \showarticletitle{Uncertainty and exploration.}
\newblock \bibinfo{journal}{\emph{Decision}} \bibinfo{volume}{6}, \bibinfo{number}{3} (\bibinfo{year}{2019}), \bibinfo{pages}{277}.
\newblock


\bibitem[\protect\citeauthoryear{Glorot, Bordes, and Bengio}{Glorot et~al\mbox{.}}{2011}]%
        {glorot2011deep}
\bibfield{author}{\bibinfo{person}{Xavier Glorot}, \bibinfo{person}{Antoine Bordes}, {and} \bibinfo{person}{Yoshua Bengio}.} \bibinfo{year}{2011}\natexlab{}.
\newblock \showarticletitle{Deep sparse rectifier neural networks}. In \bibinfo{booktitle}{\emph{Proceedings of the fourteenth international conference on artificial intelligence and statistics}}. JMLR Workshop and Conference Proceedings, \bibinfo{pages}{315--323}.
\newblock


\bibitem[\protect\citeauthoryear{Haarnoja, Zhou, Abbeel, and Levine}{Haarnoja et~al\mbox{.}}{2018a}]%
        {pmlr-v80-haarnoja18b}
\bibfield{author}{\bibinfo{person}{Tuomas Haarnoja}, \bibinfo{person}{Aurick Zhou}, \bibinfo{person}{Pieter Abbeel}, {and} \bibinfo{person}{Sergey Levine}.} \bibinfo{year}{2018}\natexlab{a}.
\newblock \showarticletitle{Soft Actor-Critic: Off-Policy Maximum Entropy Deep Reinforcement Learning with a Stochastic Actor}. In \bibinfo{booktitle}{\emph{Proceedings of the 35th International Conference on Machine Learning}} \emph{(\bibinfo{series}{Proceedings of Machine Learning Research}, Vol.~\bibinfo{volume}{80})}, \bibfield{editor}{\bibinfo{person}{Jennifer Dy} {and} \bibinfo{person}{Andreas Krause}} (Eds.). \bibinfo{publisher}{PMLR}, \bibinfo{pages}{1861--1870}.
\newblock


\bibitem[\protect\citeauthoryear{Haarnoja, Zhou, Hartikainen, Tucker, Ha, Tan, Kumar, Zhu, Gupta, Abbeel, et~al\mbox{.}}{Haarnoja et~al\mbox{.}}{2018b}]%
        {haarnoja2018soft}
\bibfield{author}{\bibinfo{person}{Tuomas Haarnoja}, \bibinfo{person}{Aurick Zhou}, \bibinfo{person}{Kristian Hartikainen}, \bibinfo{person}{George Tucker}, \bibinfo{person}{Sehoon Ha}, \bibinfo{person}{Jie Tan}, \bibinfo{person}{Vikash Kumar}, \bibinfo{person}{Henry Zhu}, \bibinfo{person}{Abhishek Gupta}, \bibinfo{person}{Pieter Abbeel}, {et~al\mbox{.}}} \bibinfo{year}{2018}\natexlab{b}.
\newblock \showarticletitle{Soft actor-critic algorithms and applications}.
\newblock \bibinfo{journal}{\emph{arXiv preprint arXiv:1812.05905}} (\bibinfo{year}{2018}).
\newblock


\bibitem[\protect\citeauthoryear{Henderson, Islam, Bachman, Pineau, Precup, and Meger}{Henderson et~al\mbox{.}}{2018}]%
        {henderson2018deep}
\bibfield{author}{\bibinfo{person}{Peter Henderson}, \bibinfo{person}{Riashat Islam}, \bibinfo{person}{Philip Bachman}, \bibinfo{person}{Joelle Pineau}, \bibinfo{person}{Doina Precup}, {and} \bibinfo{person}{David Meger}.} \bibinfo{year}{2018}\natexlab{}.
\newblock \showarticletitle{Deep reinforcement learning that matters}. In \bibinfo{booktitle}{\emph{Proceedings of the AAAI conference on artificial intelligence}}, Vol.~\bibinfo{volume}{32}.
\newblock


\bibitem[\protect\citeauthoryear{Houthooft, Chen, Duan, Schulman, De~Turck, and Abbeel}{Houthooft et~al\mbox{.}}{2016}]%
        {houthooft2016vime}
\bibfield{author}{\bibinfo{person}{Rein Houthooft}, \bibinfo{person}{Xi Chen}, \bibinfo{person}{Yan Duan}, \bibinfo{person}{John Schulman}, \bibinfo{person}{Filip De~Turck}, {and} \bibinfo{person}{Pieter Abbeel}.} \bibinfo{year}{2016}\natexlab{}.
\newblock \showarticletitle{VIME: Variational Information Maximizing Exploration}.
\newblock \bibinfo{journal}{\emph{Advances in Neural Information Processing Systems}}  \bibinfo{volume}{29} (\bibinfo{year}{2016}), \bibinfo{pages}{1109--1117}.
\newblock


\bibitem[\protect\citeauthoryear{Janner, Fu, Zhang, and Levine}{Janner et~al\mbox{.}}{2019}]%
        {janner2019trust}
\bibfield{author}{\bibinfo{person}{Michael Janner}, \bibinfo{person}{Justin Fu}, \bibinfo{person}{Marvin Zhang}, {and} \bibinfo{person}{Sergey Levine}.} \bibinfo{year}{2019}\natexlab{}.
\newblock \showarticletitle{When to trust your model: Model-based policy optimization}.
\newblock \bibinfo{journal}{\emph{Advances in Neural Information Processing Systems}}  \bibinfo{volume}{32} (\bibinfo{year}{2019}).
\newblock


\bibitem[\protect\citeauthoryear{Kidd and Hayden}{Kidd and Hayden}{2015}]%
        {kidd2015psychology}
\bibfield{author}{\bibinfo{person}{Celeste Kidd} {and} \bibinfo{person}{Benjamin~Y Hayden}.} \bibinfo{year}{2015}\natexlab{}.
\newblock \showarticletitle{The psychology and neuroscience of curiosity}.
\newblock \bibinfo{journal}{\emph{Neuron}} \bibinfo{volume}{88}, \bibinfo{number}{3} (\bibinfo{year}{2015}), \bibinfo{pages}{449--460}.
\newblock


\bibitem[\protect\citeauthoryear{Kingma and Ba}{Kingma and Ba}{2014}]%
        {kingma2014adam}
\bibfield{author}{\bibinfo{person}{Diederik~P Kingma} {and} \bibinfo{person}{Jimmy Ba}.} \bibinfo{year}{2014}\natexlab{}.
\newblock \showarticletitle{Adam: A method for stochastic optimization}.
\newblock \bibinfo{journal}{\emph{arXiv preprint arXiv:1412.6980}} (\bibinfo{year}{2014}).
\newblock


\bibitem[\protect\citeauthoryear{Kuznetsov, Shvechikov, Grishin, and Vetrov}{Kuznetsov et~al\mbox{.}}{2020}]%
        {kuznetsov2020controlling}
\bibfield{author}{\bibinfo{person}{Arsenii Kuznetsov}, \bibinfo{person}{Pavel Shvechikov}, \bibinfo{person}{Alexander Grishin}, {and} \bibinfo{person}{Dmitry Vetrov}.} \bibinfo{year}{2020}\natexlab{}.
\newblock \showarticletitle{Controlling overestimation bias with truncated mixture of continuous distributional quantile critics}. In \bibinfo{booktitle}{\emph{International Conference on Machine Learning}}. PMLR, \bibinfo{pages}{5556--5566}.
\newblock


\bibitem[\protect\citeauthoryear{Kuznetsov and Filchenkov}{Kuznetsov and Filchenkov}{2021}]%
        {kuznetsov2021solving}
\bibfield{author}{\bibinfo{person}{Igor Kuznetsov} {and} \bibinfo{person}{Andrey Filchenkov}.} \bibinfo{year}{2021}\natexlab{}.
\newblock \showarticletitle{Solving Continuous Control with Episodic Memory}. In \bibinfo{booktitle}{\emph{Proceedings of the Thirtieth International Joint Conference on Artificial Intelligence, {IJCAI-21}}}. \bibinfo{pages}{2651--2657}.
\newblock
\newblock
\shownote{Main Track.}


\bibitem[\protect\citeauthoryear{Lillicrap, Hunt, Pritzel, Heess, Erez, Tassa, Silver, and Wierstra}{Lillicrap et~al\mbox{.}}{2015}]%
        {lillicrap2015continuous}
\bibfield{author}{\bibinfo{person}{Timothy~P Lillicrap}, \bibinfo{person}{Jonathan~J Hunt}, \bibinfo{person}{Alexander Pritzel}, \bibinfo{person}{Nicolas Heess}, \bibinfo{person}{Tom Erez}, \bibinfo{person}{Yuval Tassa}, \bibinfo{person}{David Silver}, {and} \bibinfo{person}{Daan Wierstra}.} \bibinfo{year}{2015}\natexlab{}.
\newblock \showarticletitle{Continuous control with deep reinforcement learning}.
\newblock \bibinfo{journal}{\emph{arXiv preprint arXiv:1509.02971}} (\bibinfo{year}{2015}).
\newblock


\bibitem[\protect\citeauthoryear{Lin}{Lin}{1992}]%
        {lin1992self}
\bibfield{author}{\bibinfo{person}{Long-Ji Lin}.} \bibinfo{year}{1992}\natexlab{}.
\newblock \showarticletitle{Self-improving reactive agents based on reinforcement learning, planning and teaching}.
\newblock \bibinfo{journal}{\emph{Machine learning}} \bibinfo{volume}{8}, \bibinfo{number}{3-4} (\bibinfo{year}{1992}), \bibinfo{pages}{293--321}.
\newblock


\bibitem[\protect\citeauthoryear{Little and Sommer}{Little and Sommer}{2013}]%
        {little2013learning}
\bibfield{author}{\bibinfo{person}{Daniel Ying-Jeh Little} {and} \bibinfo{person}{Friedrich~Tobias Sommer}.} \bibinfo{year}{2013}\natexlab{}.
\newblock \showarticletitle{Learning and exploration in action-perception loops}.
\newblock \bibinfo{journal}{\emph{Frontiers in neural circuits}}  \bibinfo{volume}{7} (\bibinfo{year}{2013}), \bibinfo{pages}{37}.
\newblock


\bibitem[\protect\citeauthoryear{Mazoure, Doan, Durand, Pineau, and Hjelm}{Mazoure et~al\mbox{.}}{2020}]%
        {mazoure2020leveraging}
\bibfield{author}{\bibinfo{person}{Bogdan Mazoure}, \bibinfo{person}{Thang Doan}, \bibinfo{person}{Audrey Durand}, \bibinfo{person}{Joelle Pineau}, {and} \bibinfo{person}{R~Devon Hjelm}.} \bibinfo{year}{2020}\natexlab{}.
\newblock \showarticletitle{Leveraging exploration in off-policy algorithms via normalizing flows}. In \bibinfo{booktitle}{\emph{Conference on Robot Learning}}. PMLR, \bibinfo{pages}{430--444}.
\newblock


\bibitem[\protect\citeauthoryear{Mendonca, Rybkin, Daniilidis, Hafner, and Pathak}{Mendonca et~al\mbox{.}}{2021}]%
        {mendonca2021discovering}
\bibfield{author}{\bibinfo{person}{Russell Mendonca}, \bibinfo{person}{Oleh Rybkin}, \bibinfo{person}{Kostas Daniilidis}, \bibinfo{person}{Danijar Hafner}, {and} \bibinfo{person}{Deepak Pathak}.} \bibinfo{year}{2021}\natexlab{}.
\newblock \showarticletitle{Discovering and achieving goals via world models}.
\newblock \bibinfo{journal}{\emph{Advances in Neural Information Processing Systems}}  \bibinfo{volume}{34} (\bibinfo{year}{2021}).
\newblock


\bibitem[\protect\citeauthoryear{Mobin, Arnemann, and Sommer}{Mobin et~al\mbox{.}}{2014}]%
        {mobin2014information}
\bibfield{author}{\bibinfo{person}{Shariq~A Mobin}, \bibinfo{person}{James~A Arnemann}, {and} \bibinfo{person}{Fritz Sommer}.} \bibinfo{year}{2014}\natexlab{}.
\newblock \showarticletitle{Information-based learning by agents in unbounded state spaces}.
\newblock \bibinfo{journal}{\emph{Advances in Neural Information Processing Systems}}  \bibinfo{volume}{27} (\bibinfo{year}{2014}), \bibinfo{pages}{3023--3031}.
\newblock


\bibitem[\protect\citeauthoryear{Moore}{Moore}{1990}]%
        {moore1990efficient}
\bibfield{author}{\bibinfo{person}{Andrew~William Moore}.} \bibinfo{year}{1990}\natexlab{}.
\newblock \showarticletitle{Efficient memory-based learning for robot control}.
\newblock  (\bibinfo{year}{1990}).
\newblock


\bibitem[\protect\citeauthoryear{Nair, Pong, Dalal, Bahl, Lin, and Levine}{Nair et~al\mbox{.}}{2018}]%
        {nair2018visual}
\bibfield{author}{\bibinfo{person}{Ashvin Nair}, \bibinfo{person}{Vitchyr Pong}, \bibinfo{person}{Murtaza Dalal}, \bibinfo{person}{Shikhar Bahl}, \bibinfo{person}{Steven Lin}, {and} \bibinfo{person}{Sergey Levine}.} \bibinfo{year}{2018}\natexlab{}.
\newblock \showarticletitle{Visual reinforcement learning with imagined goals}. In \bibinfo{booktitle}{\emph{Proceedings of the 32nd International Conference on Neural Information Processing Systems}}. \bibinfo{pages}{9209--9220}.
\newblock


\bibitem[\protect\citeauthoryear{Pathak, Agrawal, Efros, and Darrell}{Pathak et~al\mbox{.}}{2017}]%
        {pathak2017curiosity}
\bibfield{author}{\bibinfo{person}{Deepak Pathak}, \bibinfo{person}{Pulkit Agrawal}, \bibinfo{person}{Alexei~A Efros}, {and} \bibinfo{person}{Trevor Darrell}.} \bibinfo{year}{2017}\natexlab{}.
\newblock \showarticletitle{Curiosity-driven exploration by self-supervised prediction}. In \bibinfo{booktitle}{\emph{International conference on machine learning}}. PMLR, \bibinfo{pages}{2778--2787}.
\newblock


\bibitem[\protect\citeauthoryear{Pathak, Gandhi, and Gupta}{Pathak et~al\mbox{.}}{2019}]%
        {pathak2019self}
\bibfield{author}{\bibinfo{person}{Deepak Pathak}, \bibinfo{person}{Dhiraj Gandhi}, {and} \bibinfo{person}{Abhinav Gupta}.} \bibinfo{year}{2019}\natexlab{}.
\newblock \showarticletitle{Self-supervised exploration via disagreement}. In \bibinfo{booktitle}{\emph{International conference on machine learning}}. PMLR, \bibinfo{pages}{5062--5071}.
\newblock


\bibitem[\protect\citeauthoryear{Schmidhuber}{Schmidhuber}{1991}]%
        {schmidhuber1991curious}
\bibfield{author}{\bibinfo{person}{J{\"u}rgen Schmidhuber}.} \bibinfo{year}{1991}\natexlab{}.
\newblock \showarticletitle{Curious model-building control systems}. In \bibinfo{booktitle}{\emph{Proc. international joint conference on neural networks}}. \bibinfo{pages}{1458--1463}.
\newblock


\bibitem[\protect\citeauthoryear{Silver, Lever, Heess, Degris, Wierstra, and Riedmiller}{Silver et~al\mbox{.}}{2014}]%
        {silver2014deterministic}
\bibfield{author}{\bibinfo{person}{David Silver}, \bibinfo{person}{Guy Lever}, \bibinfo{person}{Nicolas Heess}, \bibinfo{person}{Thomas Degris}, \bibinfo{person}{Daan Wierstra}, {and} \bibinfo{person}{Martin Riedmiller}.} \bibinfo{year}{2014}\natexlab{}.
\newblock \showarticletitle{Deterministic policy gradient algorithms}. In \bibinfo{booktitle}{\emph{International conference on machine learning}}. PMLR, \bibinfo{pages}{387--395}.
\newblock


\bibitem[\protect\citeauthoryear{Sukhbaatar, Lin, Kostrikov, Synnaeve, Szlam, and Fergus}{Sukhbaatar et~al\mbox{.}}{2018}]%
        {sukhbaatar2018intrinsic}
\bibfield{author}{\bibinfo{person}{Sainbayar Sukhbaatar}, \bibinfo{person}{Zeming Lin}, \bibinfo{person}{Ilya Kostrikov}, \bibinfo{person}{Gabriel Synnaeve}, \bibinfo{person}{Arthur Szlam}, {and} \bibinfo{person}{Rob Fergus}.} \bibinfo{year}{2018}\natexlab{}.
\newblock \showarticletitle{Intrinsic Motivation and Automatic Curricula via Asymmetric Self-Play}. In \bibinfo{booktitle}{\emph{International Conference on Learning Representations}}.
\newblock


\bibitem[\protect\citeauthoryear{Sutton}{Sutton}{1990}]%
        {sutton1990integrated}
\bibfield{author}{\bibinfo{person}{Richard~S Sutton}.} \bibinfo{year}{1990}\natexlab{}.
\newblock \showarticletitle{Integrated architectures for learning, planning, and reacting based on approximating dynamic programming}.
\newblock In \bibinfo{booktitle}{\emph{Machine learning proceedings 1990}}. \bibinfo{publisher}{Elsevier}, \bibinfo{pages}{216--224}.
\newblock


\bibitem[\protect\citeauthoryear{Tassa, Doron, Muldal, Erez, Li, Casas, Budden, Abdolmaleki, Merel, Lefrancq, et~al\mbox{.}}{Tassa et~al\mbox{.}}{2018}]%
        {tassa2018deepmind}
\bibfield{author}{\bibinfo{person}{Yuval Tassa}, \bibinfo{person}{Yotam Doron}, \bibinfo{person}{Alistair Muldal}, \bibinfo{person}{Tom Erez}, \bibinfo{person}{Yazhe Li}, \bibinfo{person}{Diego de~Las Casas}, \bibinfo{person}{David Budden}, \bibinfo{person}{Abbas Abdolmaleki}, \bibinfo{person}{Josh Merel}, \bibinfo{person}{Andrew Lefrancq}, {et~al\mbox{.}}} \bibinfo{year}{2018}\natexlab{}.
\newblock \showarticletitle{Deepmind control suite}.
\newblock \bibinfo{journal}{\emph{arXiv preprint arXiv:1801.00690}} (\bibinfo{year}{2018}).
\newblock


\bibitem[\protect\citeauthoryear{Thrun}{Thrun}{1992}]%
        {thrun1992efficient}
\bibfield{author}{\bibinfo{person}{Sebastian~B Thrun}.} \bibinfo{year}{1992}\natexlab{}.
\newblock \showarticletitle{Efficient exploration in reinforcement learning}.
\newblock  (\bibinfo{year}{1992}).
\newblock


\bibitem[\protect\citeauthoryear{Tully, Wells, and Morrison}{Tully et~al\mbox{.}}{2017}]%
        {tully2017exploration}
\bibfield{author}{\bibinfo{person}{Sarah Tully}, \bibinfo{person}{Adrian Wells}, {and} \bibinfo{person}{Anthony~P Morrison}.} \bibinfo{year}{2017}\natexlab{}.
\newblock \showarticletitle{An exploration of the relationship between use of safety-seeking behaviours and psychosis: A systematic review and meta-analysis}.
\newblock \bibinfo{journal}{\emph{Clinical psychology \& psychotherapy}} \bibinfo{volume}{24}, \bibinfo{number}{6} (\bibinfo{year}{2017}), \bibinfo{pages}{1384--1405}.
\newblock


\bibitem[\protect\citeauthoryear{Van~Hasselt, Guez, and Silver}{Van~Hasselt et~al\mbox{.}}{2016}]%
        {van2016deep}
\bibfield{author}{\bibinfo{person}{Hado Van~Hasselt}, \bibinfo{person}{Arthur Guez}, {and} \bibinfo{person}{David Silver}.} \bibinfo{year}{2016}\natexlab{}.
\newblock \showarticletitle{Deep reinforcement learning with double q-learning}. In \bibinfo{booktitle}{\emph{Proceedings of the AAAI conference on artificial intelligence}}, Vol.~\bibinfo{volume}{30}.
\newblock


\bibitem[\protect\citeauthoryear{van Hoof, Tanneberg, and Peters}{van Hoof et~al\mbox{.}}{2017}]%
        {van2017generalized}
\bibfield{author}{\bibinfo{person}{Herke van Hoof}, \bibinfo{person}{Daniel Tanneberg}, {and} \bibinfo{person}{Jan Peters}.} \bibinfo{year}{2017}\natexlab{}.
\newblock \showarticletitle{Generalized exploration in policy search}.
\newblock \bibinfo{journal}{\emph{Machine Learning}} \bibinfo{volume}{106}, \bibinfo{number}{9} (\bibinfo{year}{2017}), \bibinfo{pages}{1705--1724}.
\newblock


\bibitem[\protect\citeauthoryear{Vigorito}{Vigorito}{2016}]%
        {vigorito2016intrinsically}
\bibfield{author}{\bibinfo{person}{Christopher~M Vigorito}.} \bibinfo{year}{2016}\natexlab{}.
\newblock \showarticletitle{Intrinsically Motivated Exploration in Hierarchical Reinforcement Learning}.
\newblock  (\bibinfo{year}{2016}).
\newblock


\bibitem[\protect\citeauthoryear{Watkins and Dayan}{Watkins and Dayan}{1992}]%
        {watkins1992q}
\bibfield{author}{\bibinfo{person}{Christopher~JCH Watkins} {and} \bibinfo{person}{Peter Dayan}.} \bibinfo{year}{1992}\natexlab{}.
\newblock \showarticletitle{Q-learning}.
\newblock \bibinfo{journal}{\emph{Machine learning}} \bibinfo{volume}{8}, \bibinfo{number}{3-4} (\bibinfo{year}{1992}), \bibinfo{pages}{279--292}.
\newblock


\bibitem[\protect\citeauthoryear{Williams}{Williams}{1992}]%
        {williams1992simple}
\bibfield{author}{\bibinfo{person}{Ronald~J Williams}.} \bibinfo{year}{1992}\natexlab{}.
\newblock \showarticletitle{Simple statistical gradient-following algorithms for connectionist reinforcement learning}.
\newblock \bibinfo{journal}{\emph{Machine learning}} \bibinfo{volume}{8}, \bibinfo{number}{3} (\bibinfo{year}{1992}), \bibinfo{pages}{229--256}.
\newblock


\bibitem[\protect\citeauthoryear{Zhang and Van~Hoof}{Zhang and Van~Hoof}{2021}]%
        {pmlr-v139-zhang21t}
\bibfield{author}{\bibinfo{person}{Yijie Zhang} {and} \bibinfo{person}{Herke Van~Hoof}.} \bibinfo{year}{2021}\natexlab{}.
\newblock \showarticletitle{Deep Coherent Exploration for Continuous Control}. In \bibinfo{booktitle}{\emph{Proceedings of the 38th International Conference on Machine Learning}} \emph{(\bibinfo{series}{Proceedings of Machine Learning Research}, Vol.~\bibinfo{volume}{139})}, \bibfield{editor}{\bibinfo{person}{Marina Meila} {and} \bibinfo{person}{Tong Zhang}} (Eds.). \bibinfo{publisher}{PMLR}, \bibinfo{pages}{12567--12577}.
\newblock


\end{thebibliography}


\end{document}